\documentclass[lettersize,journal]{IEEEtran}

\usepackage{cite}           
\usepackage{amsmath,amsfonts,amssymb}
\usepackage{algorithmic}
\usepackage{algorithm}
\usepackage{array}
\usepackage[caption=false,font=normalsize,labelfont=sf,textfont=sf]{subfig}
\usepackage{textcomp}
\usepackage{stfloats}
\usepackage{url}
\usepackage{verbatim}
\usepackage{graphicx}
\usepackage{xcolor} 
\usepackage{soul}
\usepackage{tikz}
\usepackage{pgfplots}
\usepackage{bbm}
\usepackage{wasysym}        
\usepackage{xspace}
\usepackage{multirow}
\usepackage{booktabs}       
\usepackage{threeparttable} 
\usepackage{cuted}
\usepackage{caption}
\usepackage{adjustbox}
\usepackage{colortbl} 
\usepackage[table, svgnames]{xcolor} 

\definecolor{basecolor}{RGB}{230,240,255}  
\definecolor{bestcolor}{RGB}{220,20,60}    
\definecolor{cvprblue}{rgb}{0.21,0.49,0.74} 
\definecolor{secondcolor}{RGB}{220,127,127}

\newcommand{\ours}{IBRSteG\xspace}
\newcommand{\map}{Mapping\xspace}
\newcommand{\baseone}{Gaussian + StegaNeRF\xspace}
\newcommand{\basetwo}{Gaussian + Weng et al\xspace}
\newcommand{\base}{Vanilla Upper Bound}

\soulregister\cite7
\soulregister\ref7
\soulregister\ours7
\soulregister\base7
\soulregister\baseone7
\soulregister\basetwo7

\newcommand{\best}[1]{\textcolor{bestcolor}{\textbf{#1}}}
\newcommand{\second}[1]{\textcolor{secondcolor}{\textbf{#1}}}
\usepackage[breaklinks,colorlinks,allcolors=cvprblue,urlcolor=cvprblue]{hyperref}

\usepackage{fancyhdr}
\fancypagestyle{arxiv_copyright}{
    \fancyhf{}
    
    \fancyfoot[C]{\scriptsize \copyright~2026 IEEE. Personal use of this material is permitted. Permission from IEEE must be obtained for all other uses, in any current or future media, including reprinting/republishing this material for advertising or promotional purposes, creating new collective works, for resale or redistribution to servers or lists, or reuse of any copyrighted component of this work in other works.}
}

\begin{document}

\title{\ours: Learning a Generalizable Steganography Framework \\ for 3D Gaussian Splatting}

\author{
Fanye Kong*, 
Hongyu Xia*,
Yu Zheng, 
Boyang Gong,
Jie Zhou,~\IEEEmembership{Fellow,~IEEE}, and Jiwen Lu,~\IEEEmembership{Fellow,~IEEE}
\IEEEcompsocitemizethanks{
  \IEEEcompsocthanksitem
  The authors are with the Department of Automation, Tsinghua University, Beijing 100084, China. E-mail: kongfy23@mails.tsinghua.edu.cn, yu-zheng@tsinghua.edu.cn, xiahy23@mails.tsinghua.edu.cn,jzhou@tsinghua.edu\\.cn, lujiwen@tsinghua.edu.cn. 
  Corresponding author: Yu Zheng. 
  \IEEEcompsocthanksitem $^{*}$Fanye Kong and Hongyu Xia contributed equally to this work.
}}

\markboth{IEEE Transactions on Multimedia,~Vol.~XX, No.~X, Month~202X}%
{Author \MakeLowercase{\textit{et al.}}: Generalizable 3D Gaussian Splatting Steganography}

\maketitle

\thispagestyle{arxiv_copyright}

\vspace{-80pt}

\begin{abstract}
Recent advances in deep learning have notably improved steganographic message hiding.
However, designing a generalizable steganographic approach for 3D Gaussian Splatting (3DGS) that can embed meaningful 3D scene content remains challenging.
In this paper, we propose \ours, a generalizable framework for 3DGS steganography that enables undetectable concealment of secret scenes within a steganographic scene.
Unlike existing approaches whose parameter generation is rigidly coupled with the specific scene, we formulate 3D steganography as a feed-forward 3D Gaussian embedding process that generalizes across different 3DGS scenes.
To realize this, we introduce GAS (Gaussian Attributes Steganographer), a network that learns a scene-independent embedding function by injecting the attributes of secret 3D Gaussian points into a cover scene, thereby directly
reconstructing the steganographic scenes without per-scene finetuning or optimization.
By transforming 3D Gaussian into these structured attributes, these attributes are compatible with 2D learning paradigms and benefit from their structured nature, thereby enhancing generalization to unseen 3DGS scenes.
Extensive experiments on established datasets demonstrate that \ours can effectively conceal different scenes with high visual quality, and achieves superior capacity and security. Code is available at https://github.com/LingXiang2023/IBRSteG.

\end{abstract}


\section{Introduction}
\label{sec:intro}

The rapid growth of social media and the accelerating pace of digital information exchange have intensified the demand for secure and covert communication. Steganography provides a critical security layer by hiding both the semantic content of a message and the very existence of the communication. Compared to encryption~\cite{young1996cryptovirology} which may draw suspicion due to its detectable nature, steganographic methods facilitate covert transmission by embedding secret data within ordinary media. Driven by deep neural networks, modern steganography~\cite{hayes2017generating,baluja2019hiding,guan2022deepmih,jing2021hinet,zhu2018hidden} has evolved into a data-driven field capable of high-capacity embedding while preserving the natural statistics of the carrier~\cite{zhu2018hidden}.

As digital media evolve from 2D images to immersive 3D representations, 3D Gaussian Splatting (3DGS)~\cite{kerbl20233d} has emerged as a dominant paradigm. 3DGS is an information-rich medium favored for its real-time rendering capabilities and high fidelity. By utilizing differentiable rasterization~\cite{kopanas2022neural}, it enables accurate gradient back-propagation and stable rendering of complex scenes. These 3DGS assets carry dense geometric and appearance information, making them valuable intellectual property and ideal carriers for sensitive data. However, the open transmission of these assets across public platforms increases the risk of interception, necessitating steganographic techniques specifically designed for 3DGS to ensure secret scenes remain indistinguishable from ordinary content. 

To address this challenge, existing 3DGS steganographic methods such as GS-Hider~\cite{li2024gs} and SecureGS~\cite{securegs2025} attempt to embed one 3DGS scene within another. However, they still depend on scene-specific optimization to conceal information. This dependency creates a rigid coupling between the model and a fixed set of scenes, preventing the system from adapting to unseen assets without extensive retraining. In practical applications, this restriction poses a severe security risk. If a steganographic system is limited to a predictable set of cover scenes, an adversary can more easily monitor and detect anomalies. Furthermore, the need to retrain the network for every new secret scene incurs prohibitive computational costs and introduces significant latency, rendering real-time covert communication infeasible. Therefore, developing a generalizable steganographic approach that functions across diverse 3DGS scenes is of critical importance for the field.

\begin{figure*}
    \centering
    \includegraphics[width=\linewidth]{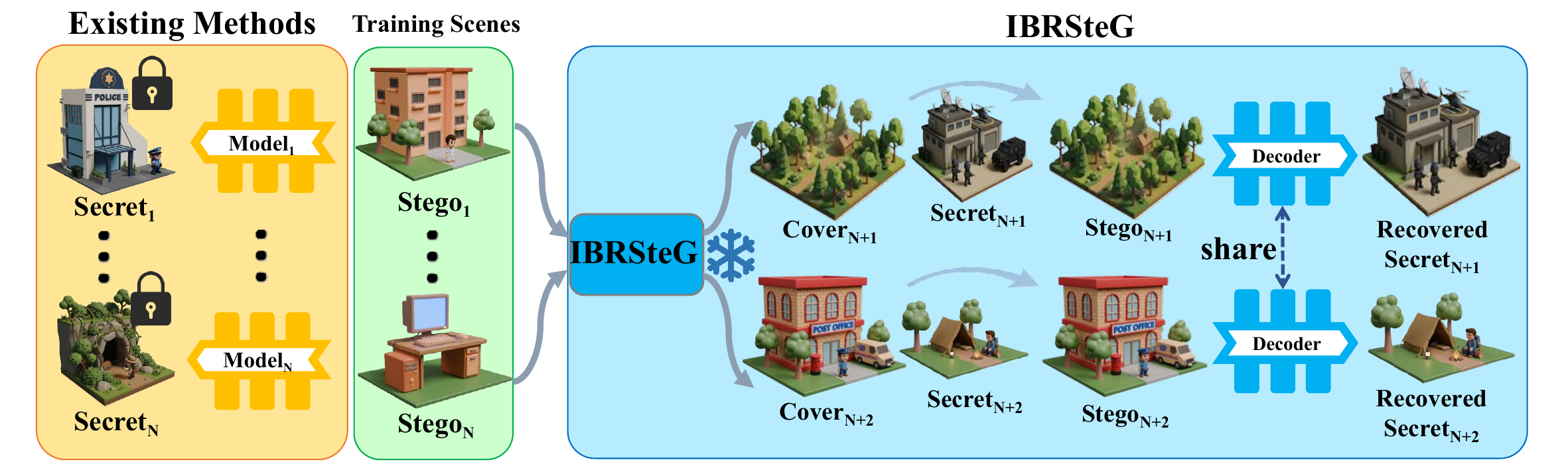}
    \vspace{-20pt} 
    \captionof{figure}{Illustration of the proposed generalizable 3D Gaussian Splatting steganography framework \ours, where a single scene-agnostic encoder–decoder embeds and extracts diverse secret 3DGS scenes within different cover scenes, in contrast to existing scene specific methods that require retraining for each scene pair.}
    \label{fig:intro}
    \vspace{-10pt} 
\end{figure*}

In this paper, we propose \ours, a generalizable framework for 3DGS steganography that achieves undetectable concealment of secret scenes. 
Unlike existing approaches whose parameter
generation is rigidly coupled with the specific scenes, we treat 3D steganography as a feed-forward embedding process that draws inspiration from Image-Based Rendering (IBR) principles to ensure cross-scene compatibility (Fig.~\ref{fig:intro}). 
To achieve this, we introduce Gaussian Attributes Steganographer (GAS), a neural network that learns a scene-agnostic embedding function. 
GAS processes the Gaussian Attribute Maps (GAM) of secret scenes and integrates them into the GAM of a cover scene. This design allows steganographic scenes to be reconstructed directly without any scene-specific fine-tuning. 
The GAM provides a structured representation of 3D Gaussian attributes, including depth, color, rotation, and opacity. By operating on these structured 2D maps, GAS directly leverages the generalization properties of 2D convolutional architectures. This strategy encourages the model to learn universal embedding patterns rather than memorizing the geometry of specific scenes. Both the embedding and extraction processes are fully differentiable, enabling end-to-end supervised training across diverse datasets. 
Extensive experiments on DyNeRF~\cite{DyNeRF}, ENeRF~\cite{ENeRF}, DTU~\cite{jensen2014large} and THuman\_MV~\cite{zhou2025gps} datasets demonstrate that \ours can effectively conceal different scenes with high visual quality.
In addition, \ours exhibits strong resilience against steganalysis, and its substantial embedding capacity indicates the potential for concealing multiple secret scenes within a single cover. We also analyze the inherent capacity-fidelity trade-off under extreme payloads. 

Our contributions can be summarized as follows:
\begin{itemize}
    \item We propose \ours , a generalizable 3DGS steganographic framework that enables the concealment of diverse secret scenes within diverse cover scenes without retraining. 
    \item We design GAS, a scene-agnostic network that learns a generalizable embedding and extraction function by operating on structured Gaussian Attribute Maps (GAM), enabling feed-forward reconstruction of steganographic scenes without per-scene finetuning.
    \item We demonstrate through extensive experiments that our method achieves superior performance in terms of visual quality, capacity, and security across multiple 3D datasets.
 
\end{itemize}


\section{Related Work}
\label{sec:related}

\subsection{3D Gaussian Splatting}
\label{sec:related_3dgs}

3D Gaussian Splatting (3DGS)~\cite{kerbl20233d} has emerged as a powerful paradigm for explicit 3D scene representation, utilizing a set of 3D anisotropic Gaussians to represent the scene geometry and appearance~\cite{wen20253d,yan2025gaussian}.
Unlike implicit representations~\cite{mildenhall2021nerf,zheng2026shadownerf,zheng2026oponerf}, 3DGS enables real-time rendering while maintaining high visual fidelity through a rasterization-based pipeline. To scale these generative capabilities to expansive environments, GaussianCity~\cite{xie2025generative} introduces a generative 3DGS framework for unbounded 3D city generation.
Recently, several works have explored generalizable 3DGS to achieve novel view synthesis across unseen scenes without per-scene optimization.
PixelSplat~\cite{charatan2024pixelsplat} and MVSGaussian~\cite{xu2024mvsgaussian} predict Gaussian primitives directly from pixel-wise features or cost volumes in a feed-forward manner.
GPS-Gaussian~\cite{zheng2024gps} introduces a depth-aware formulation to regress 3D Gaussian parameters directly from 2D image features for human novel view synthesis.
Building upon this, GPS-Gaussian+~\cite{zhou2025gps} further enhances performance by enabling robust rendering from sparse views in real-time human-background interactions.
\vspace{-10pt}

\subsection{Image Steganography}
Steganography aims to embed secret messages into cover media while maintaining perceptual indistinguishability.
Traditional methods primarily modify spatial or transform domains based on heuristic rules, including Least Significant Bit (LSB) substitution~\cite{wolfgang1996watermark}, Picture Quality Optimization~\cite{lin2010framework}, and coefficient modification in DCT~\cite{provos2003hide} or DWT~\cite{944472} domains.
However, these heuristics often introduce statistical anomalies detectable by steganalysis~\cite{lerch2016unsupervised,8727936}.
In contrast, Deep Neural Network (DNN) based approaches leverage end-to-end adversarial optimization to better preserve image statistics.
Early works utilized GANs~\cite{hayes2017generating,10306313} or encoder-decoder architectures~\cite{zhu2018hidden,baluja2019hiding} for high-capacity hiding~\cite{he2026watervib}.
Recent advances employ Invertible Neural Networks (INNs)~\cite{jing2021hinet,guan2022deepmih,li2023iscmis} to ensure lossless information flow or encode secrets into high-frequency residuals.
Concurrently, deep learning-based video steganography has also been explored~\cite{he2023adaptive,dong2022multi,yang2023centralized,10365238}, leveraging temporal coherence and redundancy for concealed message embedding.
Despite their success in the images or video domains, extending these capabilities to complex 3D representations remains a significant challenge.
\vspace{-10pt}

\subsection{Steganography for 3D Scenes}
\label{sec:3DGS stego}
With the popularity of 3D assets, steganography of 3D scene representations has attracted significant attention.
For implicit representations~\cite{mildenhall2021nerf}, existing works have explored embedding information into the network weights or color representations~\cite{li2023steganerf,luo2023copyrnerf,wang2023nerfprotector}.
Recent studies further migrate watermarks from multi-view images to radiance fields via frequency modulation~\cite{mantlemark2026}, or improve robustness with codebook-aided NeRF signatures~\cite{nerfsignature2025}.
As for 3DGS, recent methods have focused on embedding watermarks into Gaussian Splatting for copyright protection~\cite{3dgsw2025,guardsplat2025,marksplatter2025}.
However, these watermarking methods offer limited payload capacity, which makes them insufficient for steganographic tasks that require concealing entire 3D scenes.
To address this, as a pioneering work, GS-Hider~\cite{li2024gs} embeds messages into 3DGS by leveraging coupled secured feature attributes.
Building on this, SecureGS~\cite{securegs2025} enhances security and fidelity by refining the embedding strategy.
KeySS~\cite{keyss2025} introduces an explicit key-secured mechanism for protecting 3D secrets within 3DGS, while InstantSplamp~\cite{gaussianstego2024} explores a generalizable generative 3D Gaussian steganography setting.
Despite these advances, current approaches typically lack generalizability, as they require training a scene-specific encoder for each scene.
They cannot adapt to new 3DGS assets without retraining, which severely limits their practical deployment in real-world scenarios.
\vspace{-10pt}

\section{Method}
\label{sec:method}

\subsection{Preliminary}
\label{subsec:preliminary}

3D Gaussian Splatting (3DGS) represents a 3D scene as a collection of anisotropic 3D Gaussians.
Each Gaussian is defined by a center position $\mu \in \mathbb{R}^3$ and a covariance matrix $\Sigma \in \mathbb{R}^{3 \times 3}$.
The spatial influence of a Gaussian at position $x$ is defined as:
\begin{equation}
    G(x) = \exp\left(-\frac{1}{2}(x - \mu)^\top \Sigma^{-1} (x - \mu)\right).
    \label{eq:gaussian_def}
\end{equation}
To ensure the covariance matrix $\Sigma$ remains positive semi-definite during optimization, it is decomposed into a scaling matrix $\Lambda$ and a rotation matrix $R$ via $ \Sigma = R \Lambda \Lambda^\top R^\top$.

During the rendering process, 3D Gaussians are projected onto the 2D image plane.
Given the viewing transformation matrix $W$ and the Jacobian $J$ of the affine approximation of the projective transformation, the 2D covariance matrix $\Sigma'$ is computed as $\Sigma' = J W \Sigma W^\top J^\top$.
The final pixel color $I$ is computed using differentiable $\alpha$-blending of the $N$ ordered Gaussians overlapping the pixel:
\begin{equation}
    I = \sum_{i \in \mathcal{N}} c_i \alpha_i \prod_{j=1}^{i-1} (1 - \alpha_j),
    \label{eq:rendering}
\end{equation}
where $c_i$ is the view-dependent color computed from Spherical Harmonics (SH) coefficients, and $\alpha_i$ is the opacity derived from the learned opacity parameter and the spatial probability of the 2D Gaussian.

\subsection{Problem Formulation}
\label{subsec:task}
A 3DGS steganographic system is designed to conceal a secret 3DGS scene within a cover scene while ensuring its subsequent recovery. Formally, the system $\{\mathbf{F}_E,\mathbf{F}_D\}$ is defined by two core functions:
\begin{equation}
    \begin{aligned}
        &\text{Encode:} \quad S = \mathbf{F}_E(C, m), \\
        &\text{Decode:} \quad R = \mathbf{F}_D(S),
    \end{aligned}
\end{equation}
where the encoder $\mathbf{F}_E$ embeds the secret 3DGS scene $m$ into the cover 3DGS scene $C$ to produce a stego scene $S$, and the decoder $\mathbf{F}_D$ extracts the hidden 3DGS scene $R$ from $S$.

\textbf{Scene-Specific Steganography Formulation.}
Existing methods \cite{securegs2025,li2024gs} typically rely on fixed pairs of cover and secret scenes $(S_c, S_m)$ during both training and inference.
This dependence leads to scene-specific optimization that limits generalization:
\begin{equation}
\label{eq:fixed formulation}
S = \mathbf{F}_{E}^{(S_c, S_m)}(C, m), 
R = \mathbf{F}_{D}^{(S_c, S_m)}(S),
\end{equation}
where the superscript $(S_c, S_m)$ indicates that both the encoder $\mathbf{F}_E^{(S_c, S_m)}$ and decoder $\mathbf{F}_D^{(S_c, S_m)}$ are specifically optimized for the fixed cover scene $S_c$ and secret scene $S_m$, i.e.,
\begin{equation}
(\mathbf{F}_E^{(S_c, S_m)},\mathbf{F}_D^{(S_c, S_m)})
\;\neq
(\mathbf{F}_E^{(S_c', S_m')},\mathbf{F}_D^{(S_c', S_m')}).
\end{equation}
This formulation highlights the lack of generalization to unseen scenes, necessitating retraining for each new secret--cover scene pair $(S_c,S_m)$.
Such a constraint restricts the pool of viable cover scenes and compromises security, as the embedding process is confined to a small, fixed set of candidates.
Furthermore, it imposes a significant computational burden, as the model requires re-optimization upon the introduction of any new secret scene $S_m'$.

\textbf{Proposed Generalizable Steganography Formulation.}
In contrast, we propose a generalizable 3DGS steganography framework capable of operating on unseen cover--secret scene pairs $(C,m)$ without retraining.
Rather than optimizing scene-specific models, we train a unified encoder-decoder pair shared across all scenes:
\begin{equation}
\label{eq:general formulation}
\begin{aligned}
S &= \mathbf{F}_E(C, m;\,\theta_E), \\
R &= \mathbf{F}_D(S;\,\theta_D),
\end{aligned}
\end{equation}
where $(C,m)\sim\mathcal{D}$ denotes a cover--secret 3DGS scene pair sampled from a diverse scene distribution $\mathcal{D}$, and $\theta_E,\theta_D$ are scene-agnostic parameters shared across all pairs.
Only the party possessing these specific parameters can accurately extract the hidden 3DGS scene from the stego scene.

This generalizable formulation offers distinct advantages.
First, a single encoder-decoder pair $\{\mathbf{F}_E,\mathbf{F}_D\}$ operates on diverse scene pairs $(C,m)$, eliminating the need for repeated optimization and retraining per scene, thereby significantly reducing computational costs.
Second, the flexibility to employ diverse cover scenes $C\in\mathcal{D}$ enhances security, as the embedding process is not bound to a restricted, predictable set of scenes.
Finally, it facilitates practical deployment, as the system can handle unseen content $(C,m)\notin\mathcal{D}_{\text{train}}$ without retraining, making 3D steganography more scalable and reliable in real-world settings.

\begin{figure*}
    \vspace{-12pt} 
    \centering
    \includegraphics[width=\linewidth]{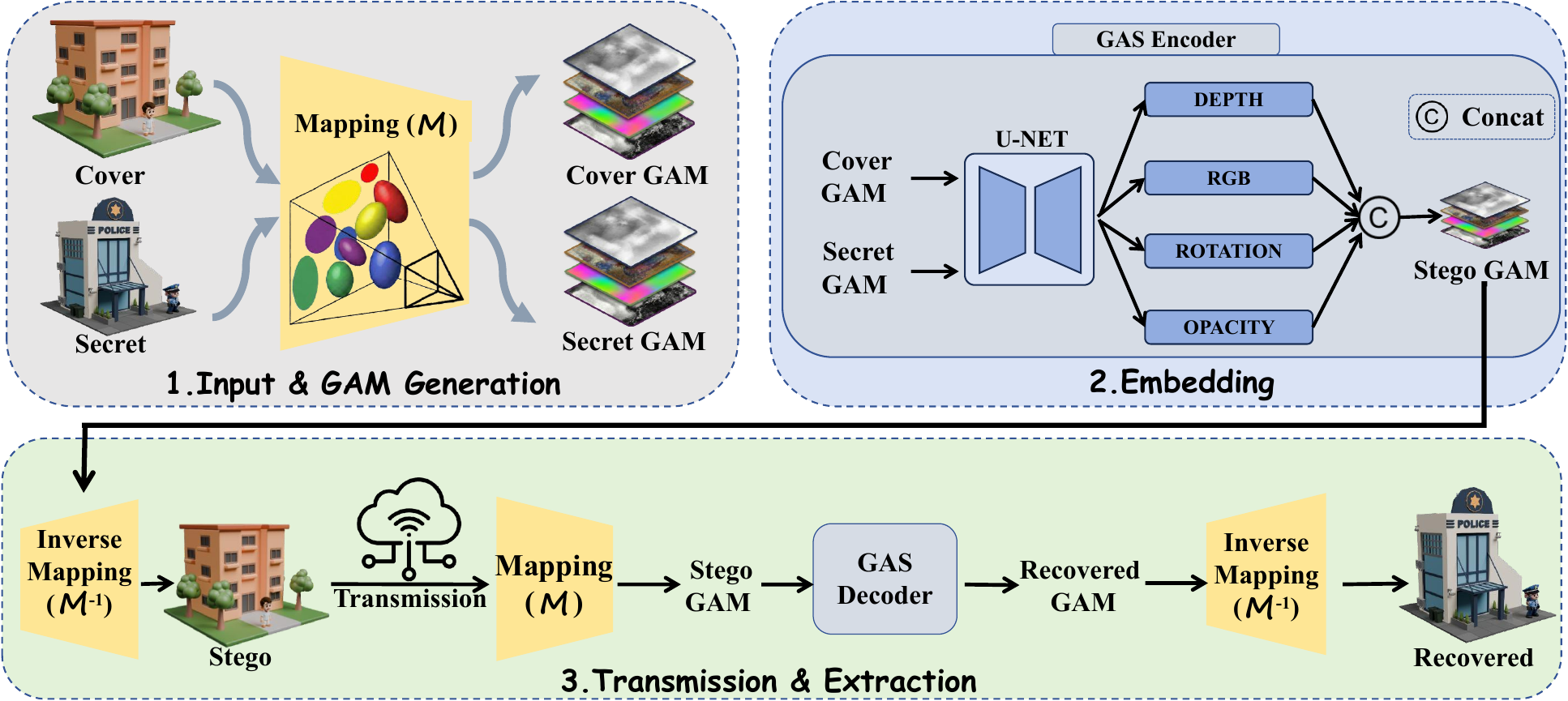}
    \captionof{figure}{Workflow of the proposed \ours framework. Top-left: Generation of the Gaussian Attribute Map (GAM) from the cover and secret scenes. Top-right: The embedding stage, where the secret GAM is injected into the cover using GAS. Bottom: The transmission and extraction process, in which the stego scene is transmitted and the secret scene is subsequently recovered.}
    \label{fig:method}
   \end{figure*}

\subsection{Generalizable 3DGS Steganography Framework}
\label{subsec:mystery}

As discussed in Sect.~\ref{subsec:task}, generalizable steganography requires a framework capable of processing unseen secret and cover scenes.
To achieve this, we propose \ours, a feed-forward generalizable framework for 3DGS steganography, as illustrated in Fig.~\ref{fig:method}.
The framework initiates by reconstructing the cover and secret scenes from dual-view images. 
These reconstructed 3DGS scenes are then mapped into Gaussian Attribute Maps (GAM) via a mapping function $M$, which establishes a structured 2D attribute representation for each scene.
These GAMs are compatible with 2D learning paradigms and leverage their structured nature to enhance generalization across unseen scenes.
On top of this representation, we introduce Gaussian Attributes Steganographer (GAS), which operates directly in the GAM domain to perform both the embedding of secret scenes into cover scenes and the subsequent extraction of the hidden secret 3DGS content.

\textbf{Mapping between 3D Gaussians and GAM.}
Formally, given a 3DGS scene $m$, the mapping process is defined as:
\begin{equation}
    G_m = \mathcal{M}(m)= \{D(m), C(m), R(m), O(m)\},
\end{equation}
where $\mathcal{M}(\cdot)$ is the mapping function that converts the input 3DGS scene $m$ into the corresponding GAM $G_m$.
Here, $D(m)$, $C(m)$, $R(m)$, and $O(m)$ correspond to the attribute maps encoding depth, color, rotation (represented in quaternion form), and opacity, respectively.

Specifically, the framework performs structured 3DGS scene reconstruction for both cover and secret scenes, avoiding direct manipulation of unstructured native 3DGS assets.
We employ a pretrained generalizable 3D reconstruction model~\cite{zhou2025gps} to predict GAMs from dual-view images, with all network parameters frozen during training.
From the predicted GAM, the corresponding 3D Gaussian representation is explicitly derived. 
Formally, given a pixel coordinate $x$ and its associated depth value $D(x)$ in the GAM, the 3D position of a Gaussian primitive $X$ is obtained via an inverse projection function:
\begin{equation}
    X = \Pi^{-1}(P, x, D(x)),
\end{equation}
where $P$ denotes the camera intrinsic and extrinsic parameters, and $\Pi^{-1}$ denotes the inverse projection transformation that lifts 2D coordinates into 3D space.

Unlike traditional 3DGS that produces an unordered point cloud with variable density, our framework adopts a pixel-aligned representation. Each pixel $x$ in the GAM stores the surface depth $D(x)$ of the first ray intersection and deterministically corresponds to exactly one Gaussian primitive. This grid-structured formulation avoids ambiguities caused by multi-hit rays or varying primitive counts. We denote this deterministic reconstruction from the 2D GAM to 3D Gaussian primitives as the inverse mapping $\mathcal{M}^{-1}$.

Accordingly, we define the representation mapping $\mathcal{M}$ as the process of indexing 3D Gaussian primitives back to their originating GAM, i.e., $\mathcal{M} : m \rightarrow G_m$, where each Gaussian center $\mathbf{X}$ is uniquely associated with a pixel coordinate $x$ under the camera projection $P$. This one-to-one correspondence enables seamless attribute embedding and accurate recovery of the secret 3DGS from the steganographic GAM.
From a security perspective, the camera parameters $P$ also serve as a compact geometric key. Even if the decoder output GAM were exposed, reconstructing a coherent 3D secret still requires the exact pre-shared intrinsics and extrinsics for the inverse mapping $\mathcal{M}^{-1}$.
\footnote{In practice, the camera parameters $P$ required for GAM reconstruction are pre-shared between the sender and receiver. This information is highly compact, incurring a negligible overhead of less than 1 KB.}

\textbf{Gaussian Attributes Steganographer (GAS) Network.}
Leveraging the structured GAM representation of the 3DGS scene, we introduce the Gaussian Attributes Steganographer (\textbf{GAS}).
By operating in this structured attribute domain, GAS learns a scene-agnostic embedding and extraction function that generalizes across unseen cover and secret scenes.
This module comprises an encoder and a decoder.
The encoder fuses the cover and secret GAMs to generate a steganographic GAM, which is subsequently mapped back to a 3DGS representation through the inverse mapping $\mathcal{M}^{-1}$.
The decoder extracts the secret GAM from the steganographic GAM, which is then transformed back into the original secret 3DGS scene.

The encoder of GAS accepts the cover and secret GAMs as inputs.
A U-Net backbone~\cite{ronneberger2015u} extracts multi-level features from the inputs and feeds them into four parallel multi-layer convolutional subnetworks: the Depth head, Color head, Rotation head, and Opacity head.
Specifically, a residual connection~\cite{he2016deep} is employed, adding the depth feature map of the cover GAM to the output of the Depth head.
This design incentivizes the Depth head to learn a residual refinement of the cover depth, rather than predicting depth values from scratch.
The outputs of the four heads are then combined to form the steganographic GAM.
This embedding process is formalized as:
\begin{equation}
\begin{aligned}
\text{Embedding}:G_s = \mathbf{F}_{E}(G_c, G_m), \quad S = \mathcal{M}^{-1}(G_s),
\end{aligned}
\end{equation}
where $G_c$ and $G_m$ are the Gaussian Attribute Maps (GAM) of the cover and secret scenes, respectively.
\(G_s\) denotes the output stego GAM, and $S$ denotes the stego 3DGS.
\(\mathcal{M}^{-1}\) represents the inverse mapping that reconstructs the stego 3DGS \(S\) from \(G_s\).

The decoder of GAS takes the stego GAM as input and utilizes a straightforward multi-layer convolutional neural network to output the recovered GAM.
The inverse mapping $\mathcal{M}^{-1}$ is then applied to reconstruct the secret 3DGS scene, completing the extraction process.
This secret extraction process is formalized as:
\begin{equation}
\begin{aligned}
&G_s=\mathcal{M}(S),\\
\text{Extraction}:\quad &G_r = \mathbf{F}_{D}(G_s),\\
&R = \mathcal{M}^{-1}(G_r),
\end{aligned}
\end{equation}
where \(G_r\) is the output recovered GAM, and $R$ is the recovered 3DGS.
Both the decoder and head modules employ the ReLU activation function and instance normalization~\cite{ulyanov2016instance}.

\ours is trained in an end-to-end framework that receives multi-view posed images from diverse scenes.
This training strategy exposes the network to a wide range of geometric and appearance variations, rather than fixed pairs of cover and secret scenes.
As a result, the model learns a generalizable embedding and extraction function applicable to diverse 3DGS scenes, independent of scene-specific optimization.

\subsection{Loss Function}
\label{subsec:loss}
\textbf{3D Loss.}
Adhering to the multi-view training protocol, we supervise both the stego and secret scenes using images rendered from multiple viewpoints.
For each view, we jointly optimize appearance fidelity and geometric consistency through image-level and geometry-level supervision.
Specifically, we employ a combination of pixel-wise reconstruction loss and structural similarity loss on the rendered RGB images.

\begin{table*}[t]
\centering
\caption{Quantitative comparison on different datasets. $\uparrow$ indicates higher is better; $\downarrow$ indicates lower is better.
\textbf{\base} denotes the reconstruction quality of GPS-Gaussian+~\cite{zhou2025gps} without steganographic embedding. Best results (excluding Bound) are marked in \textbf{red}.}
\label{tab:single_scene}
\small
\begin{tabular}{@{}llccc|ccc@{}}
\toprule
\multirow{2}{*}{\textbf{Dataset}} & \multirow{2}{*}{\textbf{Method}} & 
\multicolumn{3}{c|}{\textbf{Cover/Stego image pair}} & 
\multicolumn{3}{c}{\textbf{Secret/Recovered image pair}} \\
\cmidrule(lr){3-5} \cmidrule(lr){6-8}
& & PSNR$\uparrow$ & SSIM$\uparrow$ & LPIPS$\downarrow$ & 
PSNR$\uparrow$ & SSIM$\uparrow$ & LPIPS$\downarrow$ \\
\midrule
\multirow{4}{*}{THuman\_MV~\cite{zhou2025gps}}
  & \cellcolor{basecolor}\base~\cite{zhou2025gps}  & \cellcolor{basecolor}33.05 & \cellcolor{basecolor}0.961 & \cellcolor{basecolor}0.156 & \cellcolor{basecolor}33.05 & \cellcolor{basecolor}0.961 & \cellcolor{basecolor}0.156 \\
  & \baseone~\cite{li2023steganerf}   & 30.93 & 0.958 & 
0.210 & 17.77 & 0.763 & 0.560 \\
  & \basetwo~\cite{weng2019high}  & 14.65 & 0.604 & 0.606 & 20.66 & 0.781 & 0.420 \\
  & \ours             & \best{32.98} & \best{0.960} & \best{0.171} & \best{32.40} & \best{0.958} & \best{0.180} \\
\midrule
\multirow{4}{*}{ENeRF~\cite{ENeRF}}
  & \cellcolor{basecolor}\base~\cite{zhou2025gps}  & \cellcolor{basecolor}20.93 & \cellcolor{basecolor}0.510 & \cellcolor{basecolor}0.379 & \cellcolor{basecolor}20.93 & \cellcolor{basecolor}0.510 & \cellcolor{basecolor}0.379 \\
  & \baseone~\cite{li2023steganerf}   & 16.76 & 0.338 & 0.640 & 12.96 & 0.330 & 0.706 \\
  & \basetwo~\cite{weng2019high} & 11.97 & 
0.297 & 0.668 & 14.33 & 0.376 & 0.733 \\
 & \ours             & \best{20.94} & \best{0.533} & \best{0.393} & \best{21.01} & \best{0.537} & \best{0.400} \\
\midrule
\multirow{4}{*}{DyNeRF~\cite{DyNeRF}}
  & \cellcolor{basecolor}\base~\cite{zhou2025gps}  & \cellcolor{basecolor}28.16 & \cellcolor{basecolor}0.909 & \cellcolor{basecolor}0.186 & \cellcolor{basecolor}28.16 & \cellcolor{basecolor}0.909 & \cellcolor{basecolor}0.186 \\
  & \baseone~\cite{li2023steganerf}      & 18.37 & 0.636 & 0.552 & 18.64 & 0.719 & 0.563 \\
  & \basetwo~\cite{weng2019high} & 14.54 & 0.476 & 0.630 & 15.11 & 0.400 & 0.623 \\
   & \ours             & \best{23.73} & \best{0.871} & \best{0.314} & \best{24.29} & \best{0.880} & \best{0.309} \\
\bottomrule
\vspace{-20pt}
\end{tabular}
\end{table*}

For the embedding stage, the loss is defined as:
\begin{equation}
\begin{aligned}
\mathcal{L}_{\text{emb}} = 
&\ \lambda_{\text{mae}} \cdot \ell_{1}\!\left(I^{S}_{\text{pred}}, I^{C}_{\text{gt}}\right)
+ \lambda_{\text{ssim}} \cdot \ell_{\text{SSIM}}\!\left(I^{S}_{\text{pred}}, I^{C}_{\text{gt}}\right) \\
&\ + \lambda_{\text{chamfer}} \cdot \mathcal{L}_{\text{CD}}(S),
\end{aligned}
\end{equation}
where $I^{S}_{\text{pred}}$ denotes the rendered RGB images of the steganographic 3DGS, and $I^{C}_{\text{gt}}$ denotes the corresponding ground-truth images of the cover scene.
The geometric alignment loss $\mathcal{L}_{\text{CD}}$ is formulated as the Chamfer distance between the sets of 3D Gaussian points reconstructed from the left- and right-view GAMs, enforcing spatial consistency across dual perspectives.

Similarly, for the extraction stage, the loss is defined as:
\begin{equation}
\begin{aligned}
\mathcal{L}_{\text{ex}} = 
&\ \lambda_{\text{mae}} \cdot \ell_{1}\!\left(I^{R}_{\text{pred}}, I^{m}_{\text{gt}}\right)
+ \lambda_{\text{ssim}} \cdot \ell_{\text{SSIM}}\!\left(I^{R}_{\text{pred}}, I^{m}_{\text{gt}}\right) \\
&\ + \lambda_{\text{chamfer}} \cdot \mathcal{L}_{\text{CD}}(R),
\end{aligned}
\end{equation}
where $I^{R}_{\text{pred}}$ denotes the rendered RGB images of the recovered secret 3DGS, and $I^{m}_{\text{gt}}$ denotes the ground-truth images of the secret scene.

The Chamfer distance enforces cross-view geometric consistency by minimizing the bidirectional distance between the corresponding Gaussian point sets.
The weighting coefficients $\lambda_{\text{mae}}$, $\lambda_{\text{ssim}}$, and $\lambda_{\text{chamfer}}$ modulate the relative contributions of appearance reconstruction and geometry alignment.

Finally, the 3D training objective is given by:
\begin{equation}
\mathcal{L}_{\text{3D}}  = \mathcal{L}_{\text{emb}} + \mathcal{L}_{\text{ex}}.
\end{equation}

\textbf{2D GAM Loss.}
Supplementing the rendering-based supervision, we impose a 2D loss directly within the Gaussian Attribute Map (GAM) domain to further regularize the embedding and extraction processes.
This loss promotes consistency at the attribute level. 
Specifically, for the embedding stage, we define a GAM reconstruction loss between the steganographic GAM $G_s$ and the cover GAM $G_c$. 
For the extraction stage, a similar loss is imposed between the recovered GAM $G_r$ and the secret GAM $G_m$. 
The GAM loss is a weighted sum of attribute-wise reconstruction errors over depth, color, rotation, and opacity, denoted compactly as $\mathcal{L}_{\text{GAM}}(\cdot,\cdot)$.

The resulting 2D GAM loss is defined as:
\begin{equation}
\mathcal{L}_{\text{2D}} =
\lambda_{\text{emb}} \cdot \mathcal{L}_{\text{GAM}}(G_s, G_c)
+ \lambda_{\text{ex}} \cdot \mathcal{L}_{\text{GAM}}(G_r, G_m),
\end{equation}
where $\lambda_{\text{emb}}$ and $\lambda_{\text{ex}}$ balance the contributions of the embedding and extraction stages, respectively.

\textbf{Total Objective.} 
The final training objective combines the rendering-based 3D loss and the 2D GAM loss.
Specifically, the overall loss is defined as:
\begin{equation}
\mathcal{L}_{\text{total}} = \mathcal{L}_{\text{3D}} + \lambda_{\text{2D}} \cdot \mathcal{L}_{\text{2D}}.
\end{equation}
  \vspace{-20pt}

    
    
    

\section{Experiments}
In this section, we first describe the experimental setup, including datasets, evaluation metrics, implementation details, and baseline methods used for comparison.
We then evaluate the fidelity of our methods in Sec.~\ref{sec:main result}, followed by an analysis of its high-capacity hiding performance in Sec.~\ref{sec:high capacity}.
We further examine the security of our method against potential detection and steganographic analysis in Sec.~\ref{sec:secure}, and analyze its robustness under geometric degradation in Sec.~\ref{sec:robust}.
Finally, we provide ablation studies in Sec.~\ref{sec:ablation} to validate the design choices of our approach.

\begin{figure*}[t]
    \centering
    \includegraphics[width=0.9\linewidth]{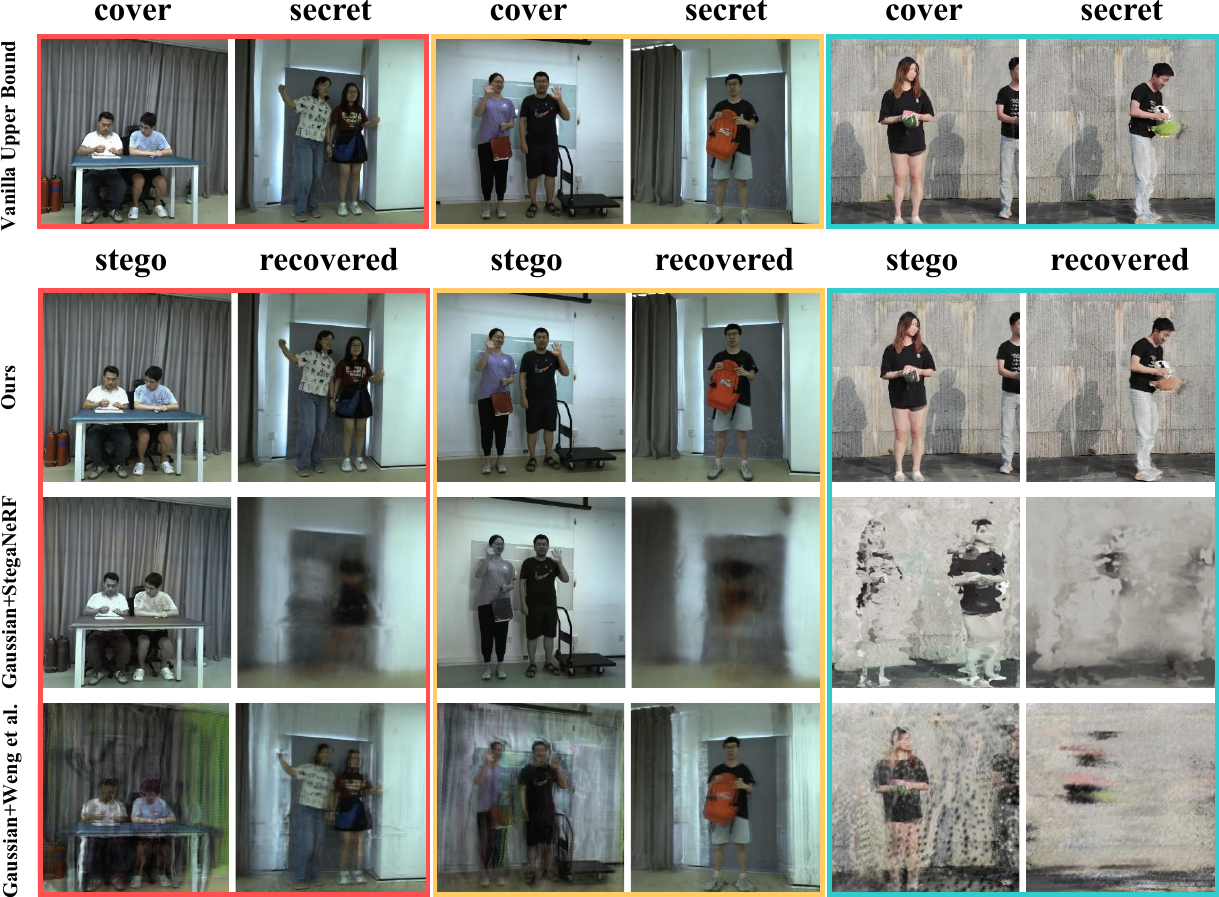} 
    \caption{Qualitative examples of stego and recovered images generated by different experimental schemes. \textbf{\base} represents the reconstruction quality of GPS-Gaussian+~\cite{zhou2025gps} without steganographic embedding.
    }
    \vspace{-5pt}
    \label{fig:main_visuals}
\end{figure*}

\begin{table*}[t]
\centering
\caption{Quantitative comparison on different secret scenes from the DTU~\cite{jensen2014large} dataset. $\uparrow$ indicates higher is better; $\downarrow$ indicates lower is better.
\textbf{\base} denotes the reconstruction quality of GPS-Gaussian+~\cite{zhou2025gps} without steganographic embedding. Best results (excluding Bound) are marked in \textbf{red}. More detailed results are provided in Appendix~A.}
\label{tab:secret_scene}
\small

\begin{tabular}{@{}llccc|ccc@{}}
\toprule
\multirow{2}{*}{\textbf{Secret Scene}} & \multirow{2}{*}{\textbf{Method}} & 
\multicolumn{3}{c|}{\textbf{Cover/Stego image pair}} & 
\multicolumn{3}{c}{\textbf{Secret/Recovered image pair}} \\
\cmidrule(lr){3-5} \cmidrule(lr){6-8}
& & PSNR$\uparrow$ & SSIM$\uparrow$ & LPIPS$\downarrow$ & 
PSNR$\uparrow$ & SSIM$\uparrow$ & LPIPS$\downarrow$ \\
\midrule
\multirow{4}{*}{Birds} 
  & \cellcolor{basecolor}\base~\cite{zhou2025gps}  & \cellcolor{basecolor}22.80 & \cellcolor{basecolor}0.724 & \cellcolor{basecolor}0.266 & \cellcolor{basecolor}29.24 & \cellcolor{basecolor}0.857 & \cellcolor{basecolor}0.239 \\
  & \baseone~\cite{li2023steganerf}   & 18.98 & 0.567 & 0.469 & 20.20 & 0.594 & 0.611 \\
  & \basetwo~\cite{weng2019high}  & 13.46 & 0.307 & 0.590 & 18.93 & 0.494 & 0.554 \\
  & \ours             & \best{22.56} & \best{0.702} & \best{0.328} & \best{23.60} & \best{0.612} & \best{0.453} \\
\midrule
\multirow{4}{*}{Bricks}
  & \cellcolor{basecolor}\base~\cite{zhou2025gps}  & \cellcolor{basecolor}25.72 & \cellcolor{basecolor}0.786 & \cellcolor{basecolor}0.232 & \cellcolor{basecolor}20.47 & \cellcolor{basecolor}0.670 & \cellcolor{basecolor}0.341 \\
  & \baseone~\cite{li2023steganerf}   & 21.67 & 0.645 & 0.443 & 14.13 & 0.423 & 0.718 \\
  & \basetwo~\cite{weng2019high} & 16.33 & 0.389 & 0.575 & 12.27 & 0.258 & 0.618 \\
  & \ours             & \best{25.40} & \best{0.766} & \best{0.296} & \best{15.99} & \best{0.435} & \best{0.522} \\
\midrule
\multirow{4}{*}{Snowman}
  & \cellcolor{basecolor}\base~\cite{zhou2025gps}  & \cellcolor{basecolor}23.13 & \cellcolor{basecolor}0.728 & \cellcolor{basecolor}0.278 & \cellcolor{basecolor}28.24 & \cellcolor{basecolor}0.844 & \cellcolor{basecolor}0.203 \\
  & \baseone~\cite{li2023steganerf}      & 19.25 & 0.566 & 0.481 & 20.63 & 0.570 & 0.596 \\
  & \basetwo~\cite{weng2019high} & 13.46 & 0.318 & 0.595 & 19.52 & 0.451 & 0.543 \\
   & \ours             & \best{22.69} & \best{0.701} & \best{0.347} & \best{22.91} & \best{0.619} & \best{0.389} \\
\midrule
\multirow{4}{*}{Tools}
  & \cellcolor{basecolor}\base~\cite{zhou2025gps}  & \cellcolor{basecolor}25.98 & \cellcolor{basecolor}0.790 & \cellcolor{basecolor}0.261 & \cellcolor{basecolor}19.68 & \cellcolor{basecolor}0.658 & \cellcolor{basecolor}0.255 \\
  & \baseone~\cite{li2023steganerf}      & 22.38 & 0.651 & 0.432 & 11.66 & 0.379 & 0.641 \\
  & \basetwo~\cite{weng2019high} & 16.34 & 0.356 & 0.588 & 12.68 & 0.349 & 0.553 \\
   & \ours             & \best{25.47} & \best{0.761} & \best{0.331} & \best{15.75} & \best{0.496} & \best{0.423} \\
\bottomrule
\vspace{-13pt}
\end{tabular}

\end{table*}

\subsection{Experimental Setups}
\label{sec:experiment setup}

\textbf{Datasets.} 
To evaluate generalization across diverse scene configurations, we adopt four widely-used datasets: DyNeRF~\cite{DyNeRF}, ENeRF~\cite{ENeRF}, THuman\_MV~\cite{zhou2025gps} and DTU~\cite{jensen2014large}, aligning with the generalizable method GPS-Gaussian+~\cite{zhou2025gps}.
DyNeRF~\cite{DyNeRF} provides four motion sequences of 300 frames each, where the first 220 frames are used for training and the remaining frames are used for testing.
ENeRF~\cite{ENeRF} provides four motion sequences for training and two sequences of unseen human-background scenes for testing.
THuman\_MV~\cite{zhou2025gps} provides 15 sequences in total, with 10 training sequences and 5 unseen testing sequences.
DTU~\cite{jensen2014large} provides a variety of non-human-centric indoor environments, from which 4 scenes are selected for testing.
All images are standardized to 1K resolution. 

\textbf{Evaluation Metrics.} 
Following existing works~\cite{DyNeRF,ENeRF,zhou2025gps,3dgsw2025}, we evaluated the experimental results using Peak Signal-to-Noise Ratio (PSNR), Structural Similarity Index (SSIM)~\cite{wang2004image}, and Learned Perceptual Image Patch Similarity (LPIPS)~\cite{zhang2018unreasonable}.
PSNR measures pixel-level differences between a reference and a processed image, while SSIM~\cite{wang2004image} evaluates perceived degradation by comparing structural information, luminance, and contrast.
LPIPS~\cite{zhang2018unreasonable} is a learned metric that quantifies perceptual similarity between image patches using calibrated deep features.

\textbf{Implementation Details.} 
We set $\lambda_\text{2D}$, $\lambda_\text{emb}$ and $\lambda_\text{ex}$ as 0.4, 1 and 0.75 respectively. 
In $Loss_\text{3D}$, we set hyperparameters $\lambda_\text{mae}$, $\lambda_\text{ssim}$, and $\lambda_\text{chamfer}$ as 0.8, 0.5 and 0.2.
For the \map module, we follow the implementation of~\cite{zhou2025gps}.
The model was trained for 100,000 iterations using the Adam optimizer~\cite{kingma2014adam} with an initial learning rate of 1e-4 and batch size of 2.
The learning rate was decayed by a factor of 0.1 every 20,000 iterations, and a weight decay of 1e-5 was applied.
All experiments were conducted on an RTX 3090 GPU.

\begin{table*}[t]
\centering

\caption{Results of performance comparisons on the THuman\_MV~\cite{zhou2025gps} dataset when embedding multiple scenes. $\uparrow$: higher is better; $\downarrow$: lower is better.
\textbf{\base} denotes the reconstruction quality of GPS-Gaussian+~\cite{zhou2025gps} without steganographic embedding. Best results (excluding Bound) are \textbf{red}.}
\vspace{-3pt}
\label{tab:multi_scene}
\small
\begin{tabular}{@{}llccc|ccc@{}}
\toprule
\multirow{2}{*}{\textbf{Capacity}} & \multirow{2}{*}{\textbf{Method}} & 
\multicolumn{3}{c|}{\textbf{Cover/Stego image pair}} & 
\multicolumn{3}{c}{\textbf{Secret/Recovered image pair}} \\
\cmidrule(lr){3-5} \cmidrule(lr){6-8}
& & PSNR$\uparrow$ & SSIM$\uparrow$ & LPIPS$\downarrow$ & 
PSNR$\uparrow$ & SSIM$\uparrow$ & LPIPS$\downarrow$ \\
\midrule
\multirow{4}{*}{2 scenes} 
  & \cellcolor{basecolor}\base~\cite{zhou2025gps}  & \cellcolor{basecolor}33.05 & \cellcolor{basecolor}0.961 & \cellcolor{basecolor}0.156 & \cellcolor{basecolor}33.05 & \cellcolor{basecolor}0.961 & \cellcolor{basecolor}0.156 \\
  & \baseone~\cite{li2023steganerf}   & 28.72 & 0.953 & 0.254 & 17.83 & 
0.766 & 0.563 \\
  & \basetwo~\cite{weng2019high}   & 19.20 & 0.797 & 0.429 & 21.65 & 0.800 & 0.413 \\
   & \ours              & \best{32.70} & \best{0.959} & \best{0.174} & \best{31.07} & \best{0.948} & \best{0.215} \\
\midrule
\multirow{4}{*}{3 scenes}
  & \cellcolor{basecolor}\base~\cite{zhou2025gps}  & \cellcolor{basecolor}33.05 & \cellcolor{basecolor}0.961 & \cellcolor{basecolor}0.156 & \cellcolor{basecolor}33.05 & \cellcolor{basecolor}0.961 & \cellcolor{basecolor}0.156 \\
  & \baseone~\cite{li2023steganerf}       & 28.01 & 0.950 & 0.267 & 16.62 & 0.743 & 0.581 \\
  & \basetwo~\cite{weng2019high}  & 18.20 & 0.772 
& 0.466 & 21.79 & 0.791 & 0.390 \\
& \ours              & \best{32.48} & \best{0.957} & \best{0.178} & \best{29.57} & \best{0.941} & \best{0.248} \\
\bottomrule
  \vspace{-5pt}
\end{tabular}
\end{table*}

\textbf{Compared Methods.} 
We conduct quantitative experiments with two groups of baselines to comprehensively evaluate the transferability of existing 3D model steganography and the migration potential of 2D steganography methods.
\textit{(1) Decoder Methods.} Most existing 3DGS and NeRF watermarking and steganographic approaches~\cite{li2023steganerf,luo2023copyrnerf,guardsplat2025,3dgsw2025,gaussianmarker2024,gaussianstego2024} employ an additional image-space decoder to recover hidden information from rendered views.
As a representative example, StegaNeRF~\cite{li2023steganerf} extracts secrets via a decoder operating on rendered images, and we denote it as ``\baseone''.
Recent 3DGS watermarking methods such as GaussianMarker~\cite{gaussianmarker2024} also follow this decoder-based paradigm, but are restricted to low-capacity, bit-level payloads and are therefore not suitable for concealing full 3D scenes.
\textit{(2) 2D Steganography Methods.} To examine whether the performance gains arise from the proposed GAS, rather than from merely operating on GAM representations, we construct a diagnostic baseline by directly migrating a representative 2D steganographic model proposed by Weng et al.~\cite{weng2019high} to the GAM domain, replacing the GAS module in our pipeline.
This baseline reflects a naive extension of 2D steganography to 3D attributes, and we denote this category as ``\basetwo''.

To ensure a fair and rigorous comparison, all baseline models are implemented with the same capacity constraints, unified decoder architectures, and identical hyperparameters and optimization schedules as our \ours framework.

\subsection{Main Results}
\label{sec:main result}

We assess the visual quality of rendered images for both the stego scene and the recovered secret scene.
Experiments are conducted on three diverse datasets, including THuman\_MV~\cite{zhou2025gps}, ENeRF~\cite{ENeRF}, and DyNeRF~\cite{DyNeRF}, to comprehensively evaluate the steganographic performance and generalizability of \ours.
We compare our method against \baseone~\cite{li2023steganerf} and \basetwo~\cite{weng2019high}.
We additionally report the reconstruction quality of GPS-Gaussian+~\cite{zhou2025gps}, the current state-of-the-art (SOTA) for generalizable 3D Gaussian Splatting, without steganographic embedding. This serves as a ``Vanilla Upper Bound"  and acts as a reference baseline to rigorously evaluate the effect of steganographic embedding on visual quality.
The quantitative results are summarized in Table.~\ref{tab:single_scene}, and the qualitative visual comparisons are presented in Fig.~\ref{fig:main_visuals}.

As shown in Table.~\ref{tab:single_scene}, \ours consistently achieves superior hiding and recovery performance across all datasets without scene-specific tuning, demonstrating strong generalizability.
On THuman\_MV~\cite{zhou2025gps}, our method attains a stego PSNR of 32.98 dB and a recovered PSNR of 32.40 dB, which are very close to the reconstruction quality obtained by directly applying GPS-Gaussian+~\cite{zhou2025gps} without steganographic embedding (33.05 dB). This small gap indicates that embedding an entire 3D scene has only a limited impact on visual quality.
In contrast, \baseone~\cite{li2023steganerf}  shows a clear imbalance between hiding and recovery. While the stego scene maintains reasonable visual quality, the recovered secret suffers a substantial degradation, with PSNR dropping to 17.77 dB.
Similarly, \basetwo~\cite{weng2019high} is constrained by the inability of 2D-based methods, leading to poor visual fidelity for both the stego and recovered scenes (e.g., Cover PSNR of 14.65 dB on THuman\_MV~\cite{zhou2025gps})
The same trends persist on the more challenging ENeRF~\cite{ENeRF} and DyNeRF ~\cite{DyNeRF} datasets. On ENeRF~\cite{ENeRF}, \ours achieves a Cover PSNR of 20.94 dB, which is comparable to the quality of direct GPS-Gaussian+~\cite{zhou2025gps} reconstruction without embedding (20.93 dB), while \baseone~\cite{li2023steganerf}  and \basetwo~\cite{weng2019high} degrades significantly. 
On DyNeRF~\cite{DyNeRF}, although a gap to direct reconstruction remains, \ours still preserves substantially higher visual quality for both cover and secret scenes.
Overall, these results indicate that \ours learns a scene-agnostic embedding function that remains robust across diverse 3DGS scenes, whereas existing methods are more sensitive to domain shifts.

\begin{figure*}[t]
    \centering
     
    \includegraphics[width=0.9\linewidth]{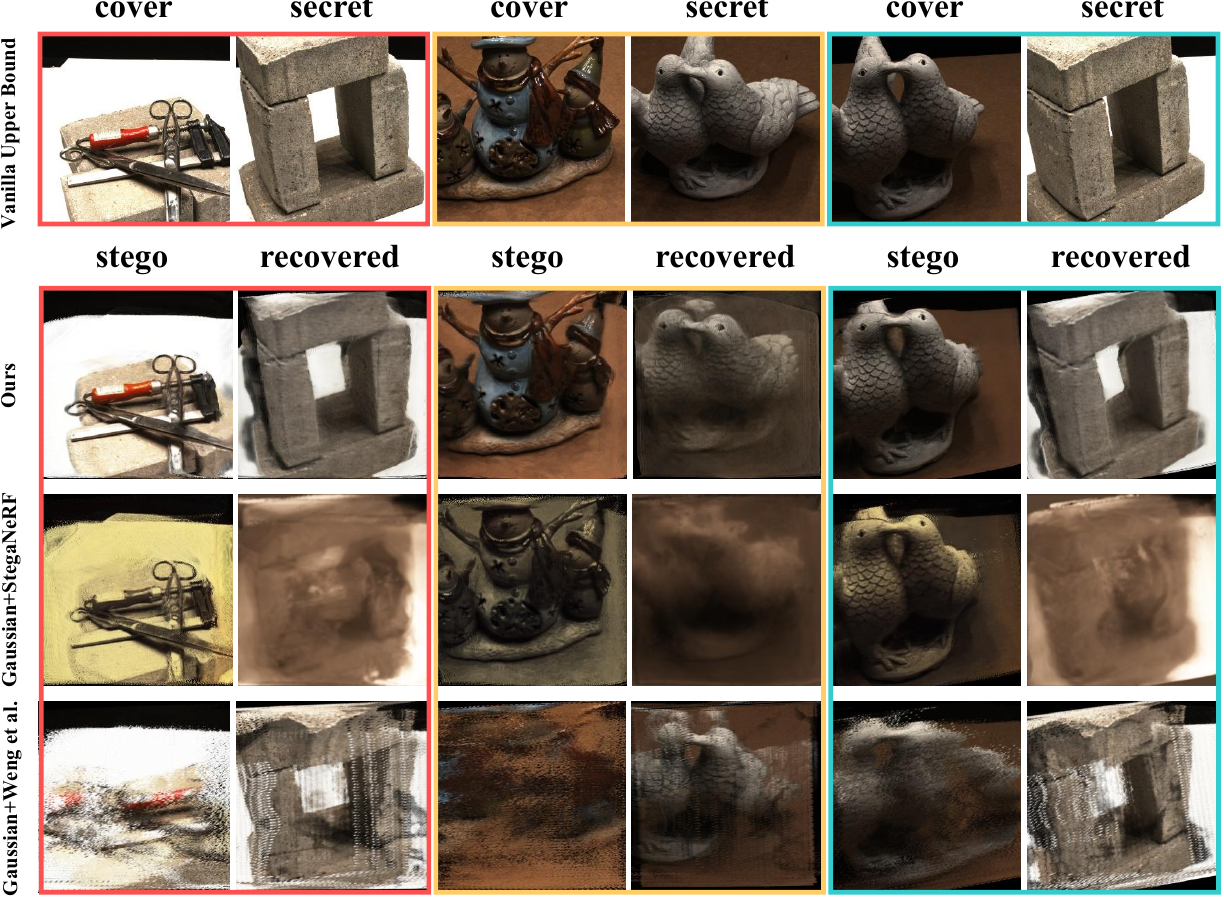} 
    \vspace{-3pt}
    \caption{Qualitative examples of stego and recovered images on the DTU~\cite{jensen2014large} dataset. \textbf{\base} represents the reconstruction quality of GPS-Gaussian+~\cite{zhou2025gps} without steganographic embedding.
    }
    \vspace{-5pt}
    \label{fig:dtu}
\end{figure*}

To further evaluate the broader applicability of our approach beyond human-centric domains, we extend our experiments to the widely used DTU~\cite{jensen2014large} dataset, which features a variety of complex, non-human-centric indoor scenes. As shown in Table~\ref{tab:secret_scene}, \ours consistently maintains high fidelity for stego scenes (e.g., 25.40 dB Cover PSNR on 'Bricks', close to upper bound) while ensuring reliable secret recovery.
In contrast, existing baselines fail to balance these objectives effectively; \baseone~\cite{li2023steganerf} suffers severe degradation in secret recovery, and \basetwo~\cite{weng2019high} experiences comprehensive quality drops across both cover and recovered images. These results further corroborate that  \ours can effectively generalize to diverse indoor environments without compromising the delicate balance between steganographic imperceptibility and recovery accuracy.

Fig.~\ref{fig:dtu} provides qualitative comparisons on the challenging DTU~\cite{jensen2014large} dataset. \ours consistently preserves complex geometries with noticeably fewer artifacts, achieving stego scenes nearly indistinguishable from the clean \base reconstructions. Conversely, \baseone~\cite{li2023steganerf} introduces noticeable blurring in recovered scenes, while \basetwo~\cite{weng2019high} exhibits severe geometric distortions. These results confirm the strong generalizability of our approach in non-human-centric environments.

\begin{figure*}[t]
    \centering
    \includegraphics[width=0.95\linewidth]{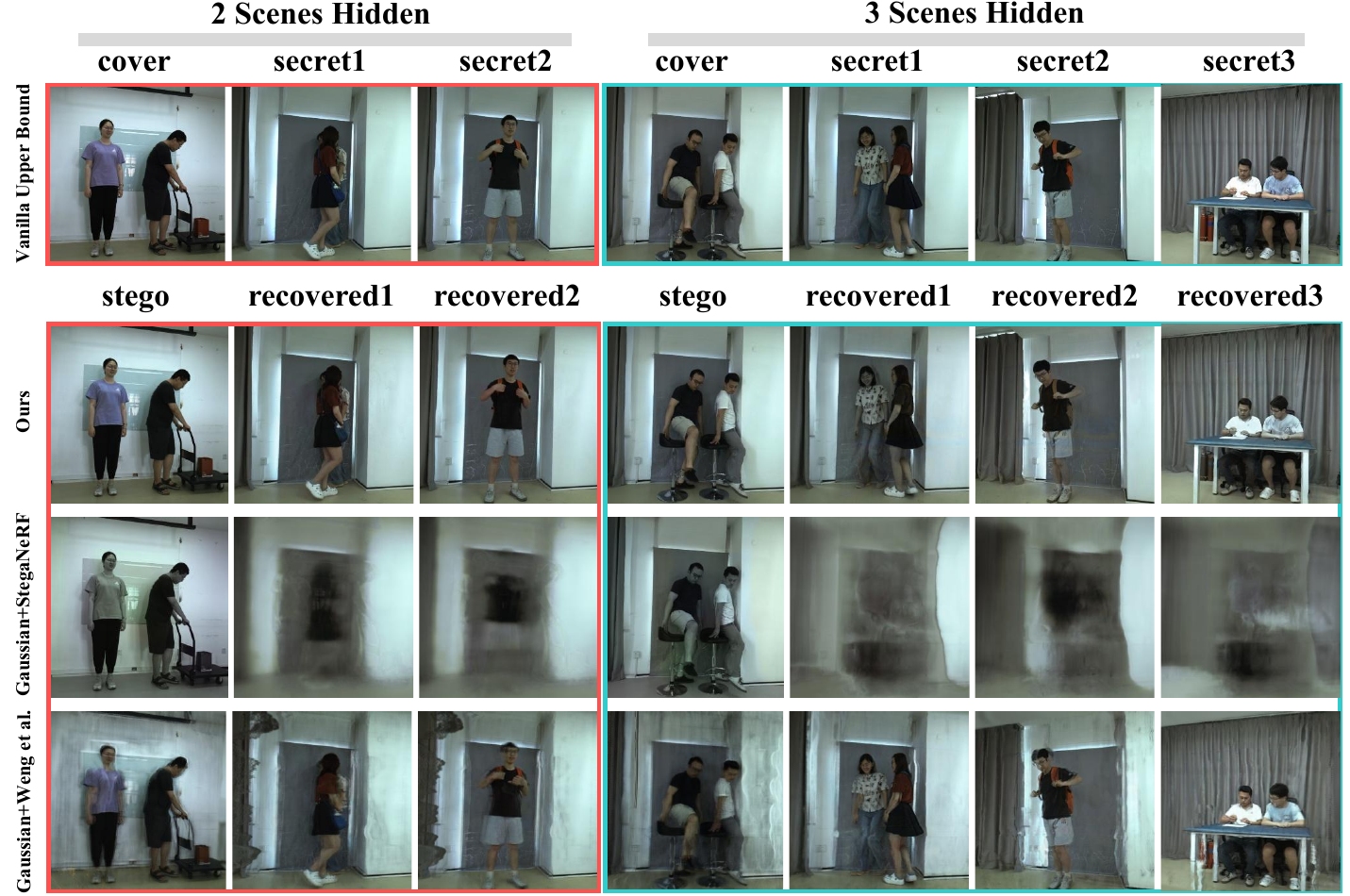}
    \vspace{-5pt}
    \caption{Qualitative examples of stego and recovered images under high-capacity embedding. \textbf{\base~\cite{zhou2025gps}} represents the reconstruction quality of GPS-Gaussian+~\cite{zhou2025gps} without steganographic embedding. }
    \label{fig:capacity_visuals}
    \vspace{-10pt}
\end{figure*}
\begin{figure}[t]
    \vspace{-5pt}
    \centering
    \includegraphics[width=0.9\linewidth]{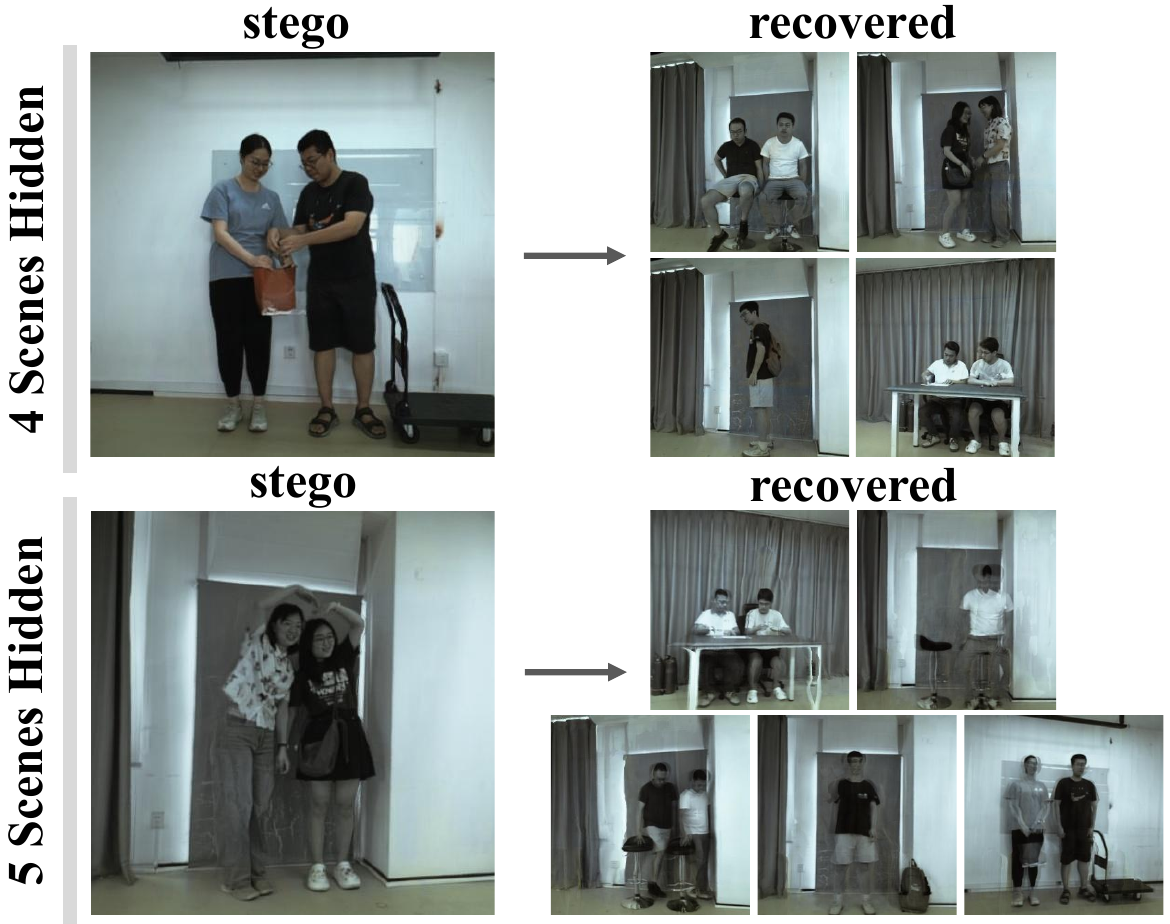}
    \vspace{-3pt}
    \caption{Visualization of stego and recovered results when embedding four and five secret scenes into a single cover scene}
    \label{fig:extreme_visuals}
    \vspace{-9pt}
\end{figure}

\subsection{Performance Under High capacity}
\label{sec:high capacity}

\noindent\textbf{Setting.}
To evaluate the payload capacity, we conduct experiments on the THuman\_MV~\cite{zhou2025gps} dataset by embedding multiple secret scenes (ranging from 2 to 5) into a single cover scene. We utilize a Multi-Channel output configuration to simultaneously recover all secret scenes. Under this setting, each hidden scene is assigned to a pre-defined segment of output channels, establishing an implicit channel indexing mechanism. This allows the decoder to deterministically disambiguate and reconstruct multiple payloads in a single feed-forward pass. We report the metrics for the stego cover scene and the average metrics of the recovered secret scenes.

\noindent\textbf{Hiding two/three secret scenes.}
We first compare our method against the baselines when embedding two and three secret scenes, with quantitative results summarized in Table.~\ref{tab:multi_scene} and visual examples shown in Fig.~\ref{fig:capacity_visuals}.
As the payload increases, \ours consistently preserves high visual quality for both the stego scene and the recovered secret scenes.
When embedding two secret scenes, \ours maintains a stego PSNR of 32.70 dB and an average recovered PSNR of 31.07 dB, indicating that increasing the payload does not significantly compromise reconstruction quality. In contrast, baselines suffer a pronounced degradation.
Similar trends are observed when embedding three secret scenes. Despite the higher payload, \ours continues to produce stable reconstructions for both cover and secret scenes.
These results suggest that the proposed \ours effectively disentangles multiple scene representations during high-capacity embedding, while conventional methods are more susceptible to interference among dense 3D features.

\noindent\textbf{Hiding $\geq 4$ secret scenes.}
To push the limits of our framework, we further increase the payload to $\geq 4$ secret scenes.
As shown in Table.~\ref{tab:high_capacity} and visualized in Fig.~\ref{fig:extreme_visuals}, \ours remains acceptable even under this extreme load.
With four secret scenes, we achieve a stego PSNR of 31.76 dB and a recovered PSNR of 28.94 dB.
When embedding five distinct secret scenes into a single cover, the stego PSNR and recovered PSNR are 28.63 dB and 24.77 dB. This performance reflects a capacity-fidelity trade-off under extreme payloads. Despite the numerical degradation, the structural integrity of the recovered scenes remains identifiable. These results demonstrate the high-capacity potential of our method while outlining its practical boundaries.


\begin{table}[t]
\centering
\caption{Image quality when embedding 4 or 5 secret scenes into one cover scene on THuman\_MV~\cite{zhou2025gps} dataset.}
\label{tab:high_capacity}
\small
\setlength{\tabcolsep}{3pt}
\begin{tabular}{@{}lccc|ccc@{}}
\toprule
\multirow{2}{*}{\textbf{Capacity}} & 
\multicolumn{3}{c|}{\textbf{Cover/Stego}} & 
\multicolumn{3}{c}{\textbf{Secret/Recovered}} \\
\cmidrule(lr){2-4} \cmidrule(lr){5-7}
& PSNR$\uparrow$ & SSIM$\uparrow$ & LPIPS$\downarrow$ & 
PSNR$\uparrow$ & SSIM$\uparrow$ & LPIPS$\downarrow$ \\
\midrule
4 scenes & 31.76 & 0.953 & 0.200 & 28.94 & 0.936 & 0.268 \\
5 scenes & 28.63 & 0.934 & 0.270 & 24.77 & 0.871 & 0.375 \\
\bottomrule
\vspace{-10pt}
\end{tabular}
\end{table}

\subsection{Steganographic Security}
\label{sec:secure}
To evaluate the security of our method against steganalysis, we conduct anti-steganography detection experiments using StegExpose~\cite{boehm2014stegexpose}, following the same protocol as prior work~\cite{jing2021hinet,li2024gs,li2024purified}.
Specifically, the detection set is constructed by mixing rendered images from the stego scenes and the corresponding non-embedded reference scenes with equal proportions.
StegExpose~\cite{boehm2014stegexpose} is then applied to this mixed set to distinguish stego images from non-stego ones.

\begin{figure}[t]
    \centering
    \includegraphics[width=0.8\linewidth]{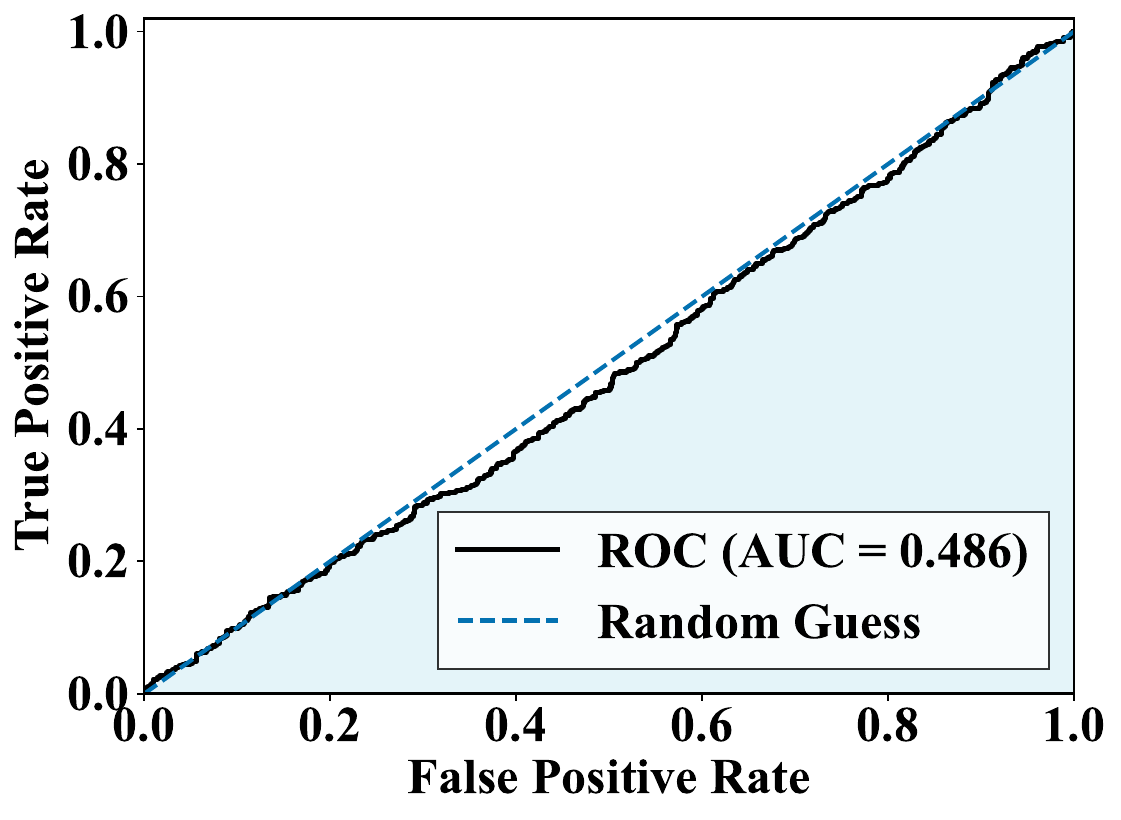}
    \caption{ROC curve of StegExpose-based steganalysis on rendered images.
    The detection set consists of an equal mixture of stego images and non-embedded reference images.
    The dashed diagonal line indicates random guessing.}
    \label{fig:stegexpose}
    \vspace{-14pt}
\end{figure}

We vary the detection threshold of StegExpose over a wide range and report the corresponding Receiver Operating Characteristic (ROC) curve in Fig.~\ref{fig:stegexpose}.
In the ideal case, a secure steganographic method should yield a detection performance close to random guessing, where the true positive rate equals the false positive rate and the Area Under the Curve (AUC) approaches 0.5.
As shown in Fig.~\ref{fig:stegexpose}, the ROC curve of our method closely follows the diagonal line, with an AUC of 0.486, indicating that StegExpose fails to reliably distinguish stego images from non-stego ones.
These results suggest that the rendered images produced by our method exhibit strong resistance to classical image-based steganalysis.

\begin{figure}[t]
    \centering
    \includegraphics[width=1.0\linewidth]{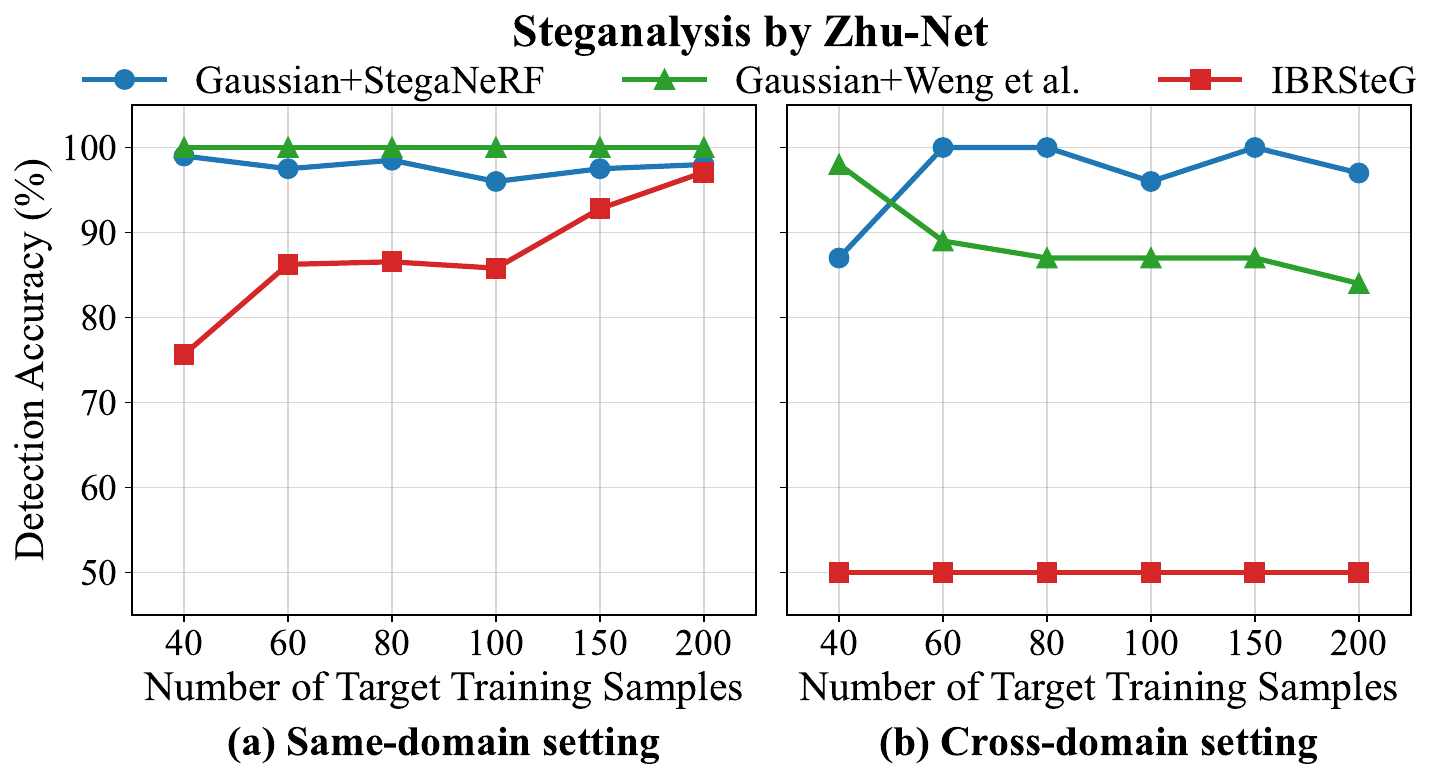}
    \vspace{-15pt}
    \caption{Detection accuracy of Zhu-Net~\cite{zhang2020depth} on stego images generated by different methods. 
    }
    \label{fig:steganalysis_comparison}
    \vspace{-2pt}
\end{figure}

Following prior work~\cite{weng2019high,jing2021hinet}, we further evaluate the undetectability of our method under learning-based steganalysis, which measures how many stego samples are required for a deep detector to reliably distinguish stego scenes from cover scenes.
Specifically, we adopt a state-of-the-art model Zhu-Net~\cite{zhang2020depth} as the steganalysis model and progressively increase the number of training image pairs available to the attacker.

We conduct two groups of experiments under different testing conditions.
In the first setting (same-domain), the steganalysis model is trained on rendered cover-stego image pairs generated from the THuman\_MV~\cite{zhou2025gps} dataset and evaluated on unseen scenes from the same dataset.
As shown in the left of Fig.~\ref{fig:steganalysis_comparison}, our method consistently exhibits lower detection accuracy compared to baseline methods.
Notably, Zhu-Net requires more than 150 image pairs to achieve reliable detection performance (above 90\%), whereas competing baselines can be detected with substantially fewer samples.
This indicates that stego images produced by our method are significantly harder to learn and classify, even when the attacker operates within the same data domain.

In the second setting (cross-domain), the steganalysis model is still trained on THuman\_MV~\cite{zhou2025gps}-generated image pairs but tested on stego images rendered from the ENeRF dataset~\cite{ENeRF}, which differs notably in scene structure and appearance.
As shown in the right of Fig.~\ref{fig:steganalysis_comparison}, the detection accuracy remains close to random guessing (around 50\%) across all training sample sizes.
This suggests that steganalysis models trained on one dataset fail to generalize to stego images from a substantially different 3D scene distribution, which shows that our method maintains strong security guarantees under realistic deployment conditions.
These results show that our method exhibits strong resistance to learning-based steganalysis.

Beyond image-based evaluation, exploring detection approaches directly in the 3D attribute domain is an important consideration, although dedicated 3DGS steganalysis methods are currently scarce. To assess security in this domain, we conduct a statistical analysis of the native 3D Gaussian attributes. Excluding the unaltered scale and numerically normalized rotation attributes, we focus on measurable deviations in RGB, depth, and opacity.

As summarized in Table~\ref{tab:enerf_dtu_transfer}, we derive optimal detection thresholds based on the mean and 95th percentile (q95) of these attributes using cover and stego scenes from the ENeRF dataset~\cite{ENeRF}. We then evaluate the accuracy of this statistical detector on the unseen DTU dataset. While attributes such as RGB and opacity exhibit numerical deviations on the source dataset, these statistical thresholds fail to generalize. When applied to the DTU~\cite{jensen2014large} dataset, the detection accuracy for most metrics collapses to random guessing. The highest cross-dataset transferability achieved is merely 60\% for the opacity q95 statistic. This severe performance drop demonstrates that such statistical differences are highly dataset-dependent rather than indicative of generalized steganographic artifacts.

Overall, our empirical results demonstrate that our proposed method exhibits strong resistance to both 2D image-based steganalysis and 3D statistical detection.

\begin{table}[bt!]
\centering
\caption{Detection accuracy on the DTU~\cite{jensen2014large} dataset using thresholds derived from the ENeRF~\cite{ENeRF} dataset.}
\vspace{-3pt}
\label{tab:enerf_dtu_transfer}
\small
\setlength{\tabcolsep}{4pt}
\begin{adjustbox}{max width=\linewidth}

\begin{tabular}{@{}lcccc@{}}
\toprule
\textbf{Criterion} & \textbf{Cover} & \textbf{ Stego} & \textbf{Threshold} & \textbf{Accuracy} \\
\midrule
rgb(mean) & $0.1620$ & $0.2567$ & $> 0.204347$ & $0.5000$ \\
rgb(q95) & $0.5525$ & $0.6370$ & $> 0.605187$ & $0.5000$ \\
depth(mean) & $0.1820$ & $0.1821$ & $> 0.178077$ & $0.5000$ \\
depth(q95) & $0.2258$ & $0.2258$ & $< 0.252990$ & $0.5000$ \\
opacity(mean) & $0.3778$ & $0.3130$ & $< 0.343775$ & $0.5000$ \\
opacity(q95) & $0.9122$ & $0.6992$ & $< 0.805869$ & $0.6042$ \\
\bottomrule
\end{tabular}

\end{adjustbox}
\vspace{-2mm}
\end{table}

\subsection{Robustness Under Geometric Degradation}
\label{sec:robust}
To assess the robustness of \ours, we experiment under geometric degradation via random pruning. Specifically, we randomly remove 5\% to 20\% of the Gaussians from the stego scene and evaluate the subsequent recovery quality. As summarized in Table~\ref{tab:robustness}, \ours exhibits resilience to this degradation; even with 20\% of the Gaussian points discarded, the secret scenes can still be recovered with acceptable visual fidelity. These results suggest that the embedded steganographic information is distributed across the Gaussian primitives rather than being concentrated in a few points, providing a degree of tolerance against random pruning.
Furthermore, we conduct additional evaluations against four other types of perturbations: resplatting, compression, quantization, and spatial cropping. \ours achieves promising robustness under these conditions as well, with comprehensive results detailed in Appendix B.

\begin{table}[bt!]
\centering
\caption{Robustness analysis under random pruning. Metrics represent the quality of the recovered secret images after discarding 5\%--20\% of the Gaussian points.}
\label{tab:robustness}
\small
\begin{tabular}{@{}lccc@{}}
\toprule
\textbf{Ratio} & PSNR$\uparrow$ & SSIM$\uparrow$ & LPIPS$\downarrow$ \\
\midrule
5\%   & \textbf{29.79} & \textbf{0.8797} & \textbf{0.3889} \\
10\%  & \textbf{28.15} & \textbf{0.8283} & \textbf{0.4479} \\
15\%  & \textbf{26.97} & \textbf{0.7851} & \textbf{0.4852} \\
20\%  & \textbf{25.91} & \textbf{0.7453} & \textbf{0.5153} \\
\bottomrule
\end{tabular}
\vspace{-5pt}
\end{table}

\subsection{Comparison with Non-Generalizable Approaches}
\label{sec:non-generalizable}
GS-Hider~\cite{li2024gs}, SecureGS~\cite{securegs2025}, and KeySS~\cite{keyss2025} are representative 3DGS~\cite{kerbl20233d} steganography methods operating under a scene-specific training paradigm. These methods necessitate time-consuming retraining or optimization for each fixed cover-secret scene pair, which fundamentally differs from the generalizable steganography setting considered in our work. To further highlight this architectural distinction, we quantitatively compare the computational cost and reconstruction quality, as reported in Table~\ref{tab:reply_dtu_thu_compare_2}. On an NVIDIA RTX 3090 GPU, training a GS-Hider~\cite{li2024gs} model for a single scene pair takes approximately 2.5 hours, while the KeySS~\cite{keyss2025} baseline requires 7.6 hours per scene. In contrast, our proposed generalizable model executes the entire end-to-end process in approximately two seconds per scene without requiring any scene-specific optimization. It is important to clarify that this two-second execution time accounts for the complete pipeline, which fully encompasses the reconstruction of Gaussian Attribute Maps (GAM) from dual-view images. Furthermore, quantitative results on the THuman\_MV~\cite{zhou2025gps} dataset indicate that \ours achieves a secret recovery PSNR of 32.40 dB and an SSIM of 0.958, outperforming the optimization-based baselines in recovery fidelity while maintaining a competitive stego SSIM of 0.960. This substantial gap in runtime, combined with the quantitative rendering quality across both image pairs, demonstrates the efficiency and effectiveness of our method for practical applications.

\begin{table}[bt!]
\centering
\caption{Comparison of image quality, runtime between scene-specific baselines and our method on the THuman\_MV~\cite{zhou2025gps} dataset.}
\vspace{-3pt}
\label{tab:reply_dtu_thu_compare_2}
\small
\setlength{\tabcolsep}{3pt}
\begin{adjustbox}{max width=\columnwidth}

\begin{tabular}{@{}lccc|ccc|c@{}}
\toprule
\multirow{2}{*}{\textbf{Method}} &
\multicolumn{3}{c|}{\textbf{Cover/Stego image pair}} &
\multicolumn{3}{c|}{\textbf{Secret/Recovered image pair}} &
\multirow{2}{*}{\textbf{Time}} \\
\cmidrule(lr){2-4} \cmidrule(lr){5-7}
& PSNR$\uparrow$ & SSIM$\uparrow$ & LPIPS$\downarrow$ & PSNR$\uparrow$ & SSIM$\uparrow$ & LPIPS$\downarrow$ & \\
\midrule
GS-Hider~\cite{li2024gs} & \best{34.04} & \second{0.942} & \second{0.204} & \second{24.05} & \second{0.884} & \second{0.281} & \second{2.5 h} \\
KeySS~\cite{keyss2025} & 16.81 & 0.581 & 0.413 & 17.17 & 0.677 & 0.421 & 7.6 h \\
\ours & \second{32.98} & \best{0.960} & \best{0.171} & \best{32.40} & \best{0.958} & \best{0.180} & \best{2 s} \\
\bottomrule
\end{tabular}

\end{adjustbox}
\vspace{-5pt}
\end{table}


\subsection{Ablation Studies on Loss Design}
We evaluate the necessity of the proposed 3D loss $\mathcal{L}_{3D}$ and 2D GAM loss $\mathcal{L}_{2D}$ in Sec.~\ref{subsec:loss} by removing each component from the full objective.
Without $\mathcal{L}_{3D}$, the model fails to converge to a reasonable solution, and the rendering quality degrades severely, with the PSNR on the THuman\_MV~\cite{zhou2025gps} dataset dropping below 20 dB.
In contrast, removing $\mathcal{L}_{2D}$ does not cause training instability but leads to a consistent performance drop, where both stego and recovered PSNR decrease by approximately 0.2 dB. 
These results show that $\mathcal{L}_{3D}$ is essential for stable convergence, while $\mathcal{L}_{2D}$ provides complementary improvements to reconstruction quality.

\subsection{Discussion}

While IBRSteG demonstrates robust generalizability across diverse scenes, it is important to acknowledge the performance constraints imposed by the existing backbone. Because our framework utilizes the frozen GPS-Gaussian+ model to generate the Gaussian Attribute Maps (GAM) and perform initial scene reconstruction, it inherently inherits the representational constraints of this  backbone. If the underlying reconstruction model underperforms on certain challenging scene types, such as large-scale outdoor environments, the degraded quality of the GAM representation will limit the fidelity of the embedding and the extraction processes.

However, the core design of our Gaussian Attributes Steganographer (GAS) maintains a relatively low dependency on the specific backbone. As more advanced and domain-versatile generalizable 3DGS reconstruction models are developed, they are expected to be integrated into the IBRSteG pipeline. Overall, our approach represents a promising direction for 3DGS steganography, with the potential for more significant progress alongside the continuous evolution of 3D scene representation.

\label{sec:ablation}

\section{Conclusion}
In this paper, we propose \ours, the first generalizable 3D Gaussian Splatting steganography framework that enables secret scene embedding without scene-specific retraining.
To this end, we introduce GAS (Gaussian Attributes Steganographer), which learns a feed-forward embedding function by injecting secret 3D Gaussian attributes into a cover scene, enabling direct reconstruction of steganographic scenes.
This is achieved by transforming 3D Gaussians into structured attributes, thereby enhancing generalization to unseen 3DGS scenes.
Extensive experiments demonstrate that our method achieves high visual quality, high-capacity embedding, and strong security across diverse 3D scenes.
Future work will explore extending \ours to large-scale and more complex 3D environments beyond human-centric scenes.
Additionally, we plan to investigate dedicated steganalysis methods to further evaluate the security of this 3DGS steganography framework.

\vspace{-10pt}
\section*{Acknowledgments}
This work was supported in part by the National Natural Science Foundation of China under Grant 62125603, Grant 62336004, Grant 62321005, Grant 62441616, and Grant 62506198, in part by the China Postdoctoral Science Foundation under Grant 2024M761674, in part by the Beijing Natural Science Foundation Undergraduate ``Qiyan" Program, and in part by the Tsinghua University Undergraduate Academic Research Advancement Program.

\bibliographystyle{IEEEtran}
\bibliography{main}


\begin{IEEEbiography}[{\includegraphics[width=1in,height=1.25in,clip,keepaspectratio]{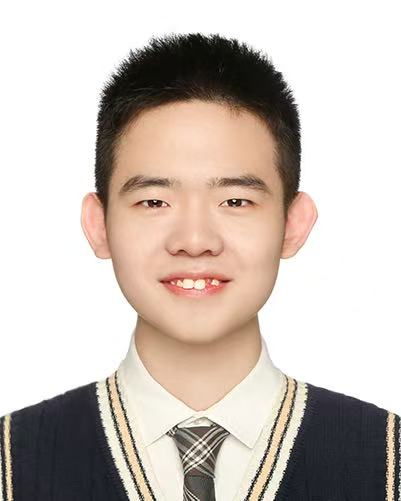}}]{Fanye Kong}
is currently pursuing the B.S. degree from the Weiyang College, Tsinghua University,
Beijing, China. He is a member of the Spark Program of Tsinghua University. His research interests include computer vision and AI safety.
\end{IEEEbiography}

\begin{IEEEbiography}[{\includegraphics[width=1in,height=1.25in,clip,keepaspectratio]{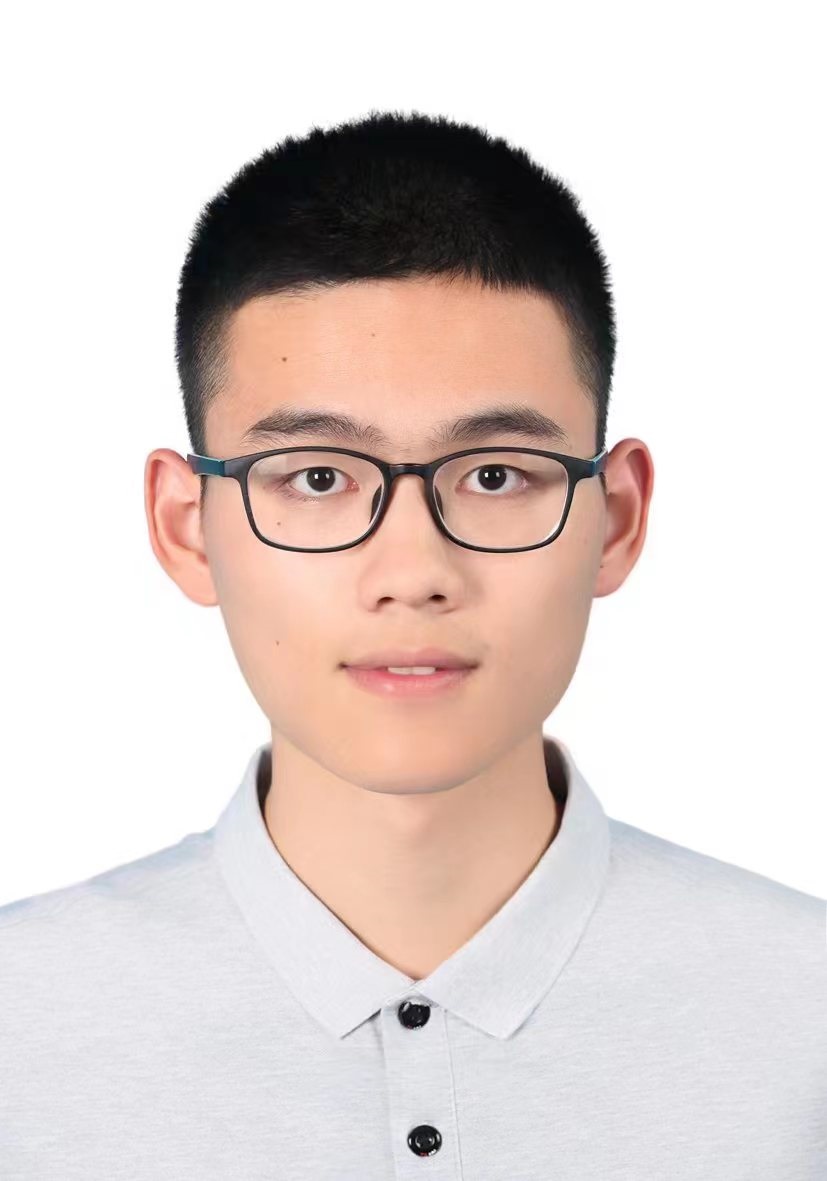}}]{Hongyu Xia}
is currently pursuing the B.S. degree in the Department of Automation at Tsinghua University, Beijing, China. His research interests include computer vision and robotic manipulation.
\end{IEEEbiography}

\begin{IEEEbiography}[{\includegraphics[width=1in,height=1.25in,clip,keepaspectratio]{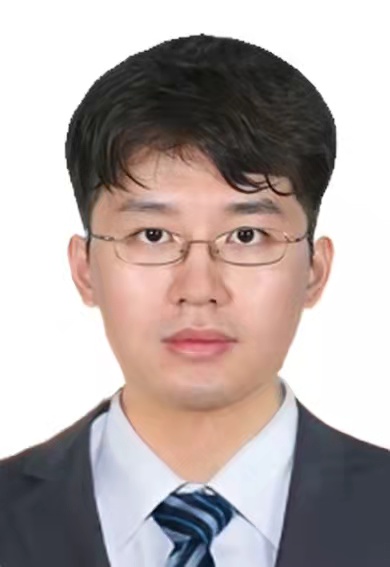}}]{Yu Zheng}
	received the BS and PhD degrees from the Department of Automation, Tsinghua University, in 2019 and 2024, respectively. 
    He is currently an assistant researcher with the Department of Automation, Tsinghua University. 
    His research interests include computer vision, AI security and pattern recognition. 
    He has published over 10 scientific papers in TPAMI, IJCV, TIP, TIFS, TMM, CVPR, ICCV, ECCV and ICLR. 
    He serves as a regular reviewer member for TPAMI, TIP, TMM, TCSVT, CVPR, ICCV, ECCV, AAAI and ICME. 
\end{IEEEbiography}

\vspace{-24pt}
\begin{IEEEbiography}[{\includegraphics[width=1in,height=1.25in,clip,keepaspectratio]{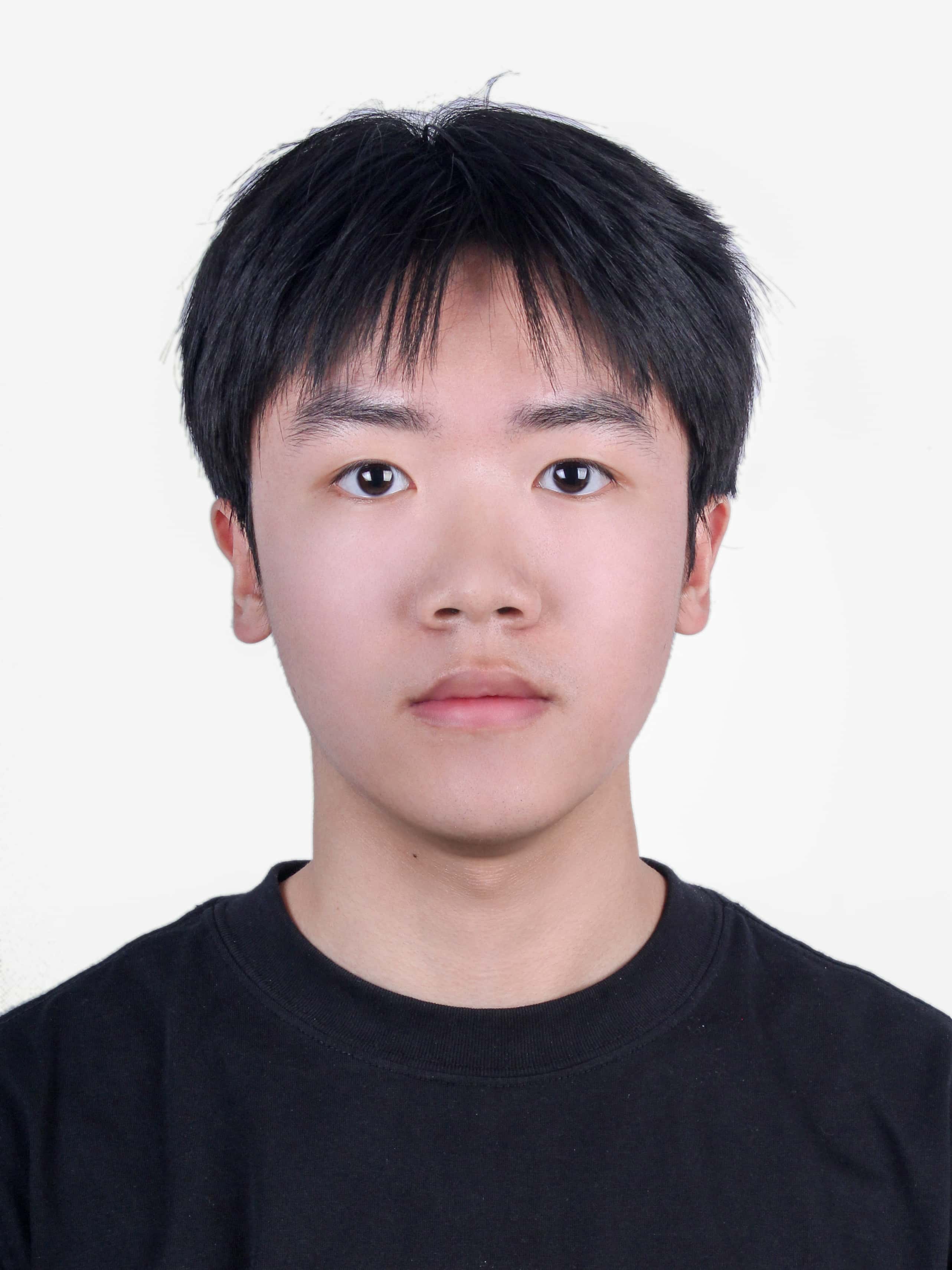}}]{Boyang Gong}
	received the BS degree from the School of Computer Science, Northwestern Polytechnical University, in 2025.
    He is currently pursuing the MS degree with the Department of Automation, Tsinghua University.
His research interests include computer vision, AI security, and generative model forensics.
\end{IEEEbiography}

\vspace{-24pt}
\begin{IEEEbiography}[{\includegraphics[width=1in,height=1.25in,clip,keepaspectratio]{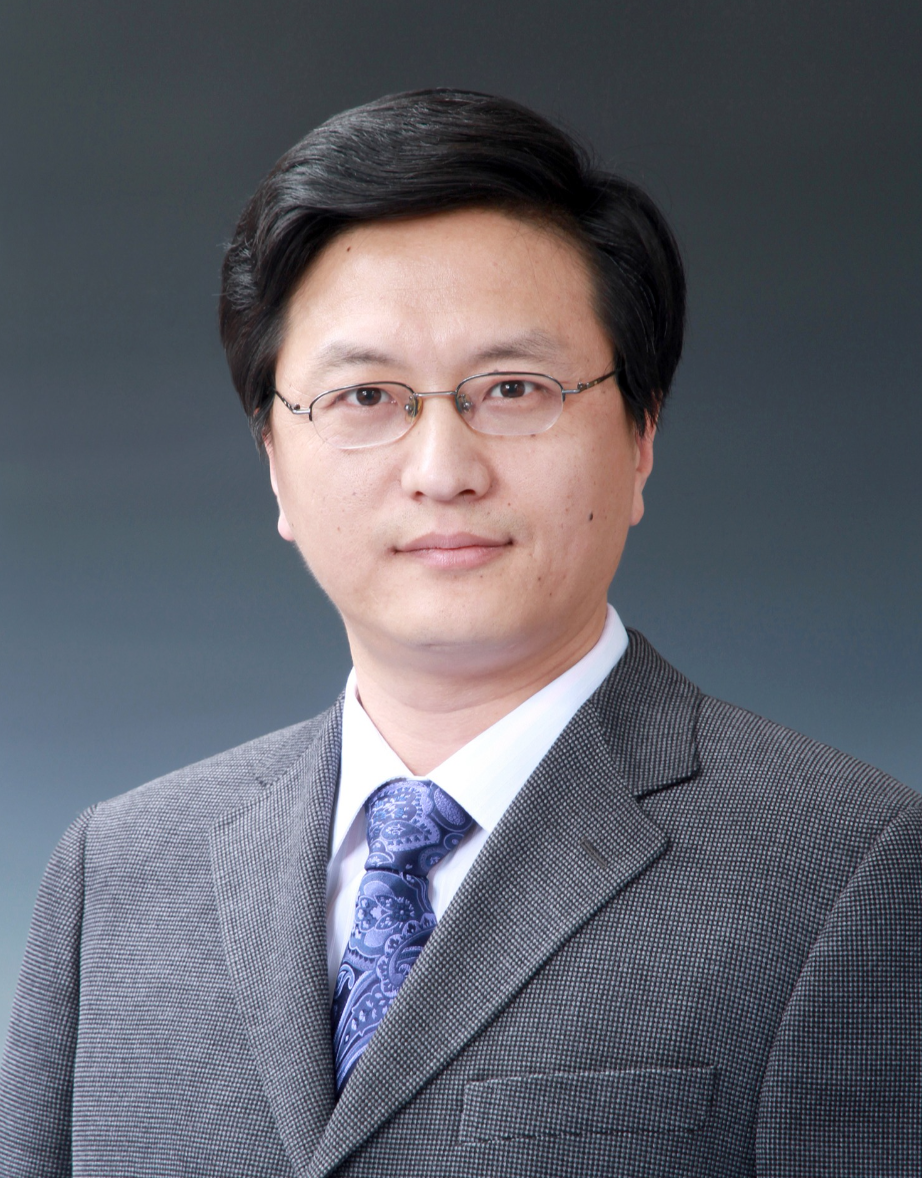}}]{Jie Zhou}
	(Fellow, IEEE) received the BS and MS degrees both from the Department of Mathematics, Nankai University, Tianjin, China, in 1990 and 1992, respectively, and the PhD degree from the Institute of Pattern Recognition and Artificial Intelligence, Huazhong University of Science and Technology (HUST), Wuhan, China, in 1995. From then to 1997, he served as a postdoctoral fellow in the Department of Automation, Tsinghua University, Beijing, China. Since 2003, he has been a full professor in the Department of Automation, Tsinghua University. His research interests include computer vision, pattern recognition, and image processing. In recent years, he has authored more than 300 papers in peer-reviewed journals and conferences. Among them, more than 100 papers have been published in top journals and conferences such as the IEEE Transactions on Pattern Analysis and Machine Intelligence, IEEE Transactions on Image Processing, and CVPR. He is an associate editor for the IEEE Transactions on Pattern Analysis and Machine Intelligence and two other journals. He received the National Outstanding Youth Foundation of China Award. He is an IEEE/IAPR Fellow. 
\end{IEEEbiography}
\vspace{-24pt}
\begin{IEEEbiography}[{\includegraphics[width=1in,height=1.25in,clip,keepaspectratio]{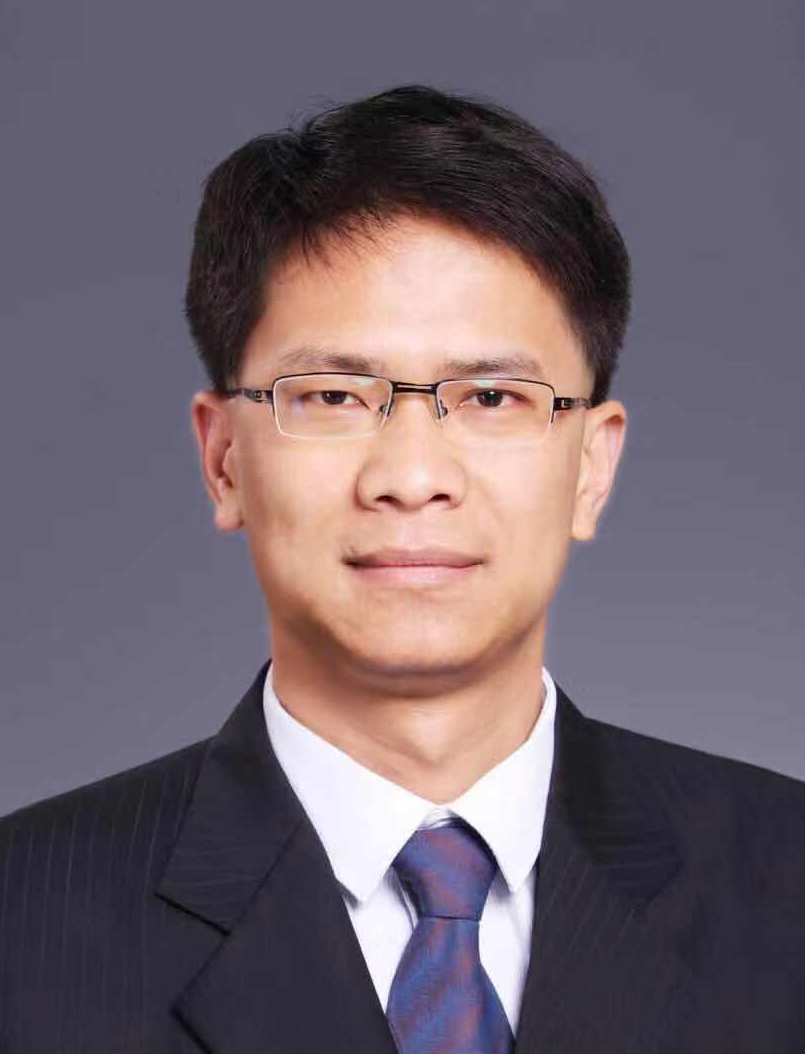}}]{Jiwen Lu}
	(Fellow, IEEE) received the B.Eng. degree in mechanical engineering and the M.Eng. degree in electrical engineering from the Xi’an University of Technology, Xi’an, China, in 2003 and 2006, respectively, and the Ph.D. degree in electrical engineering from Nanyang Technological University, Singapore, in 2012. From 2011 to 2015, He was with the Advanced Digital Sciences Center, Singapore. In November 2015, he joined the Department of Automation, Tsinghua University, where he is currently a full professor and the deputy chair of the department. His current research interests include computer vision, pattern recognition, embodied intelligent and artificial intelligence security. He serves/has served as the Co-Editor-of-Chief for Pattern Recognition Letters, an Associate Editor for the IEEE Transactions on Image Processing, the IEEE Transactions on Circuits and Systems for Video Technology, and the IEEE Transactions on Biometrics, Behavior, and Identity Sciences, and Pattern Recognition. He was a recipient of the National Natural Science Funds for Distinguished Young Scholar. He is an IEEE/IAPR Fellow. 
\end{IEEEbiography}

\end{document}